%% file: main.tex
\documentclass[11pt]{article}
\usepackage[margin=1in]{geometry}
\usepackage{graphicx}
\usepackage{subcaption}
\usepackage{transparent}
\usepackage{listings}
\usepackage{ifthen}
\usepackage[colorlinks=true,allcolors=blue]{hyperref}

\usepackage[ruled,vlined,lined,linesnumbered]{algorithm2e}

\title{Implementation and discussion of the ``Pith Estimation on Rough Log End Images using Local Fourier Spectrum Analysis'' method}
\author{%
Henry Marichal\\
\small Instituto de Ingenieria Electrica, Facultad de Ingenieria, Universidad de la Republica, Uruguay\\
\small \texttt{henry.marichal@fing.edu.uy}
\and
Diego Passarella\\
\small Sede Tacuarembo, CENUR Noreste, Universidad de la Republica, Uruguay\\
\small \texttt{diego.passarella@cut.edu.uy}
\and
Gregory Randall\\
\small Instituto de Ingenieria Electrica, Facultad de Ingenieria, Universidad de la Republica, Uruguay\\
\small \texttt{randall@fing.edu.uy}
}
\date{}
\begin{document}
\maketitle
\begin{abstract}
In this article, we analyze and propose a Python implementation of the method "Pith Estimation on Rough Log End images using Local Fourier Spectrum Analysis" \cite{Schraml2013}, by Rudolf Schraml and Andreas Uhl. The algorithm is tested over two datasets. 
\end{abstract}
\noindent\textbf{Code availability.} A CPU Python 3.11 implementation of the method accompanies this manuscript and is available at \url{https://github.com/hmarichal93/pith_detector_shraml_uhl_2013.git}. Usage instructions are included in the \verb|README.md| file of the archive.

\noindent\textbf{Keywords.} cross-section tree pith detection, tree ring delineation, Hough transform

\section{Introduction}
\begin{figure}
\begin{center}
   \begin{subfigure}{0.40\textwidth}
   \includegraphics[width=\textwidth]{figures/urudendro/F02d}
   \caption{}
   \label{fig:F02d}
   \end{subfigure}
   \begin{subfigure}{0.45\textwidth}
   \def\svgwidth{\linewidth}
   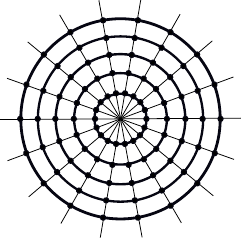
   \caption{}
   \label{fig:definitionsA}
   \end{subfigure}
   \hfill
   \caption{(a) RGB image of disk F02d from UruDendro \cite{UruDendro} dataset, (b) The whole structure, called \textit{spider web}, is formed by a \textit{center} (which corresponds to the slice pith), \textit{rays} and the \textit{rings} (concentric curves). In the scheme, the \textit{rings} are circles, but in practice, they can be (strongly) deformed as long as they don't intersect another \textit{ring}.} 
   \label{fig:main_idea}
\end{center}
\end{figure}

Pith's location on tree cross-sections is important for the forestry industry, for example, for the industrial manipulation of tree logs.  Some tree ring delineation algorithms over wood cross-section images are sensitive to a precise pith location, especially when those algorithms are based on the ring structure, a concentric pattern similar to a spider web as illustrated in Figure \ref{fig:main_idea}. Figure \ref{fig:main_idea}.a illustrates an image of a tree cross-section (or slice), and Figure \ref{fig:main_idea}.b illustrates the \textit{spider web} pattern. Ideally, the intersection point between the perpendicular lines through the tree rings should be the pith (the center of the structure, located inside the medulla of the tree). However, the \textit{spider web} model is only a general approximation. Real slices include ring asymmetries, cracks, and knots, etc. Consequently, detecting the pith position must be robust to the presence of such perturbations. 

To use the \textit{spider web} structure, many algorithms use the following steps:

\begin{enumerate}
    \item Locally estimate the normal directions to the ring structure.
    \item Compute the geometric intersection location between the lines supported by the normal directions estimated in the previous step. Sometimes, this is done using an accumulator space in a Hough Transform style.
    \item Extract a robust estimation of the pith location using the intersections defined in the previous step.
\end{enumerate}

In this article, we analyze the method "Pith Estimation on Rough Log End images using Local Fourier Spectrum Analysis" \cite{Schraml2013}, proposed by Rudolf Schraml and Andreas Uhl in 2013, and propose a CPU Python implementation of it. Their approach estimates the tree ring normals by splitting the slice image into patches and using the 2D Fourier Transform to compute the local orientation of the rings on the patch. They use an accumulation space, as in a Hough Transform approach, to search for a maximum of the intersections between the lines defined by the normals.

Other Hough-based methods have appeared since Schraml et al. \cite{Schraml2013} contribution. Kurdthongmee et al. \cite{Kurdthongmee2018} proposed the Histogram Orientation Gradient ($HOG$) to estimate the tree ring local orientations and then compute their perpendicular orientation as the normal local ring estimation. Local ring orientation is determined by the value of the highest bin of the $HOG_i$ (i indexing the image patch). Decelle et al. \cite{Decelle2022} proposed a method based on an ant colony optimization algorithm for the line accumulation step. Additionally, they proposed a region-growing approach around the global maxima to extract the pith position. They compute the mean location of all the elements of the accumulation space with a value higher than $Ts \times Mg$, where $Mg$ is the value of the global maxima and $Ts$ is a parameter between 0 and 1. Firstly, the method runs over the full image, returning a peak candidate. Then, the method is repeated in a square region centered around the first peak candidate. Gazo et al.\cite{Gazo2020} proposed the use of a weighted line accumulation step, where the weight increments linearly according to the distance from the outer edge so that the external growth rings will have less influence than the internal ones in the final result. They tested their method over computed tomography images.

Deep Neural Network (DNN) methods have also been applied to solve the pith detection problem. Kurdhongmeed et al. \cite{KURDTHONGMEE2020},  compared the effectiveness of two DNN object detector models (YoloV3 and SSD MobileNet) to locate the pith. They trained the models via transfer learning over 345  wood slice RGB images captured within a sawmill environment and used data augmentation techniques. The obtained models were evaluated over a separate dataset of 215 images. Those images present several difficulty levels, from clearly observable annual tree rings to images with annual ring features (highly) degraded by the sawing process.  The two DNN models were compared with the Kurdhongmeed et al. \cite{Kurdthongmee2018} pith detection method. The authors claim that a pre-trained SSD MobileNet got the minimum average distance error between all the evaluated methods, performing six times better than a non-DNN method. Additionally, they reported that the pith detection execution time in DNN methods is forty times lower than in non-DNN methods. 

\section{Method}
Here, we briefly describe the implementation of the approach. For further details on the method, please refer to Schraml et al. \cite{Schraml2013}.

\subsection{Pipeline}
\begin{figure}
    \begin{center}
        \begin{subfigure}{0.3\textwidth}
            \includegraphics[width=\textwidth]{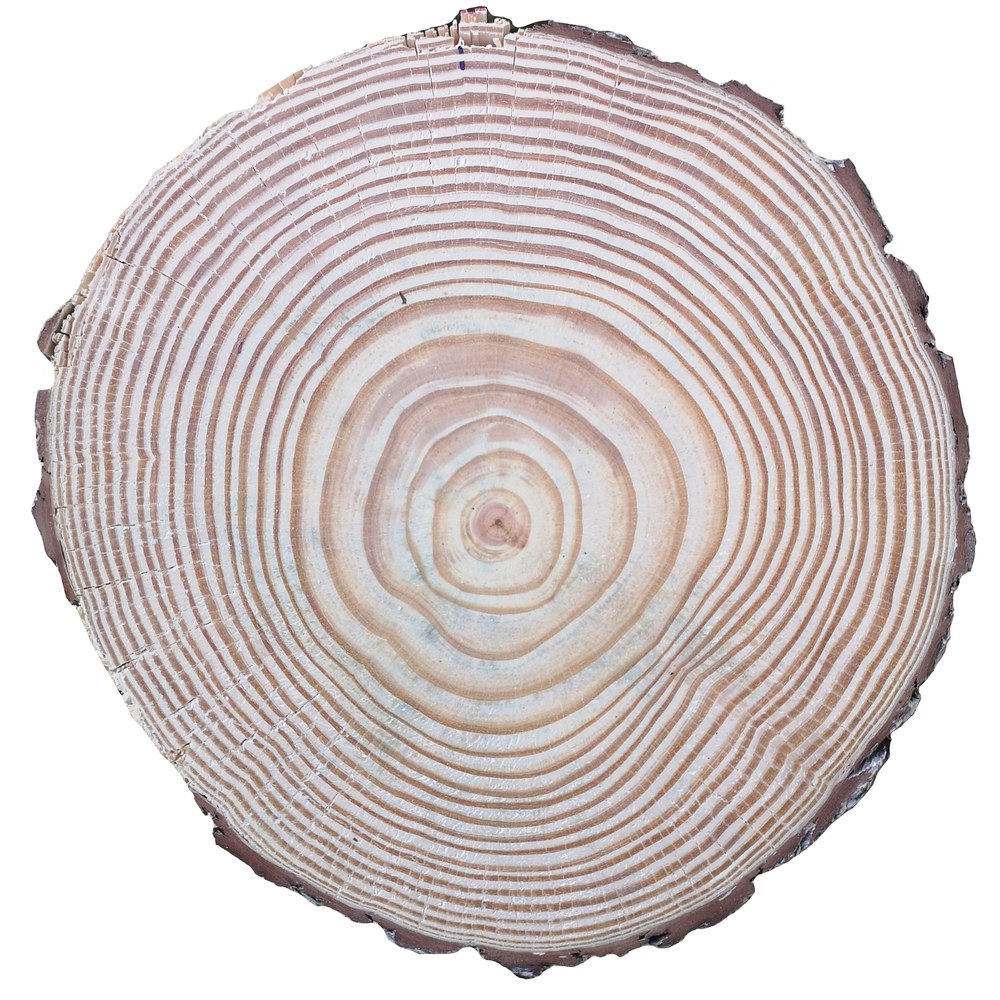}
            \caption{}
            \label{fig:img}
        \end{subfigure}
                \begin{subfigure}{0.3\textwidth}
            \includegraphics[width=\textwidth]{figures/implementation/lo/img_filtered_lines}
            \caption{}
            \label{fig:lo_lines}
        \end{subfigure}\\
        \vspace{10pt}
        \begin{subfigure}{0.3\textwidth}
            \includegraphics[width=\textwidth]{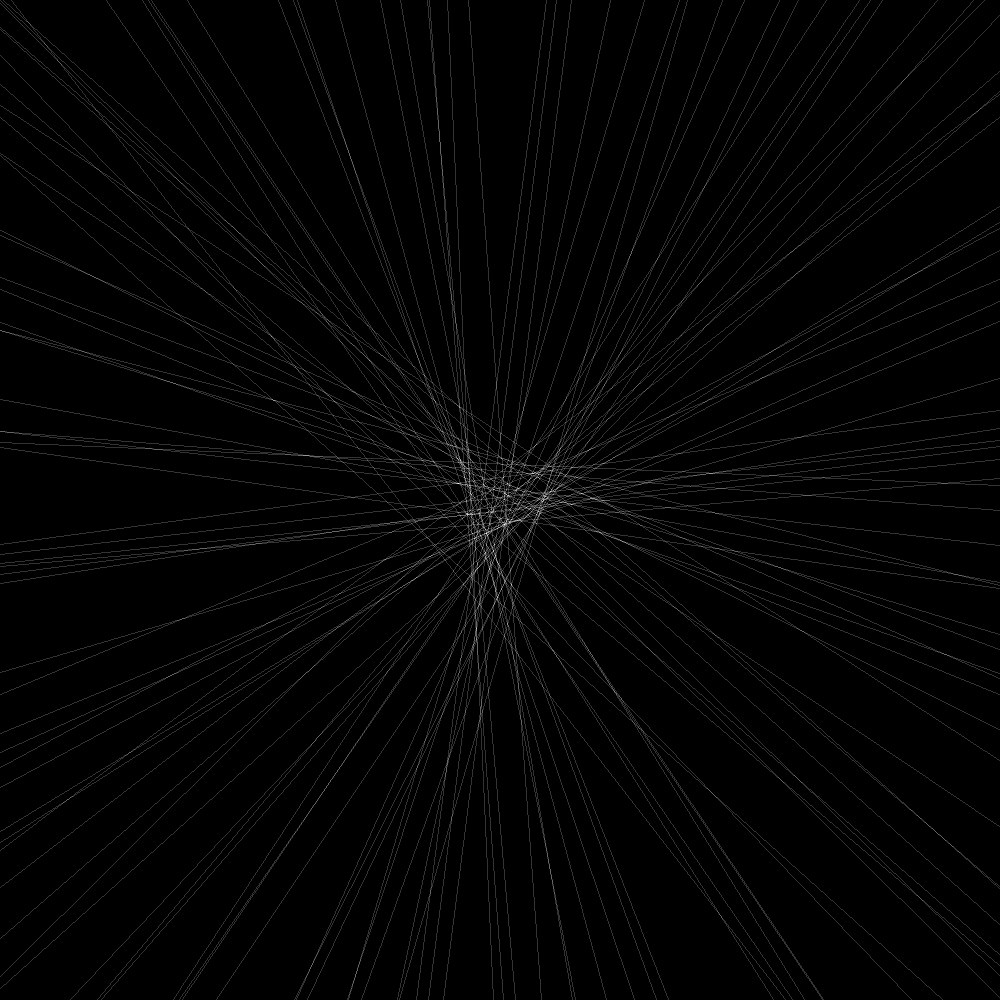}
            \caption{}
            \label{fig:accumulator}
        \end{subfigure}
        \vspace{10pt}
        \begin{subfigure}{0.3\textwidth}
            \includegraphics[width=\textwidth]{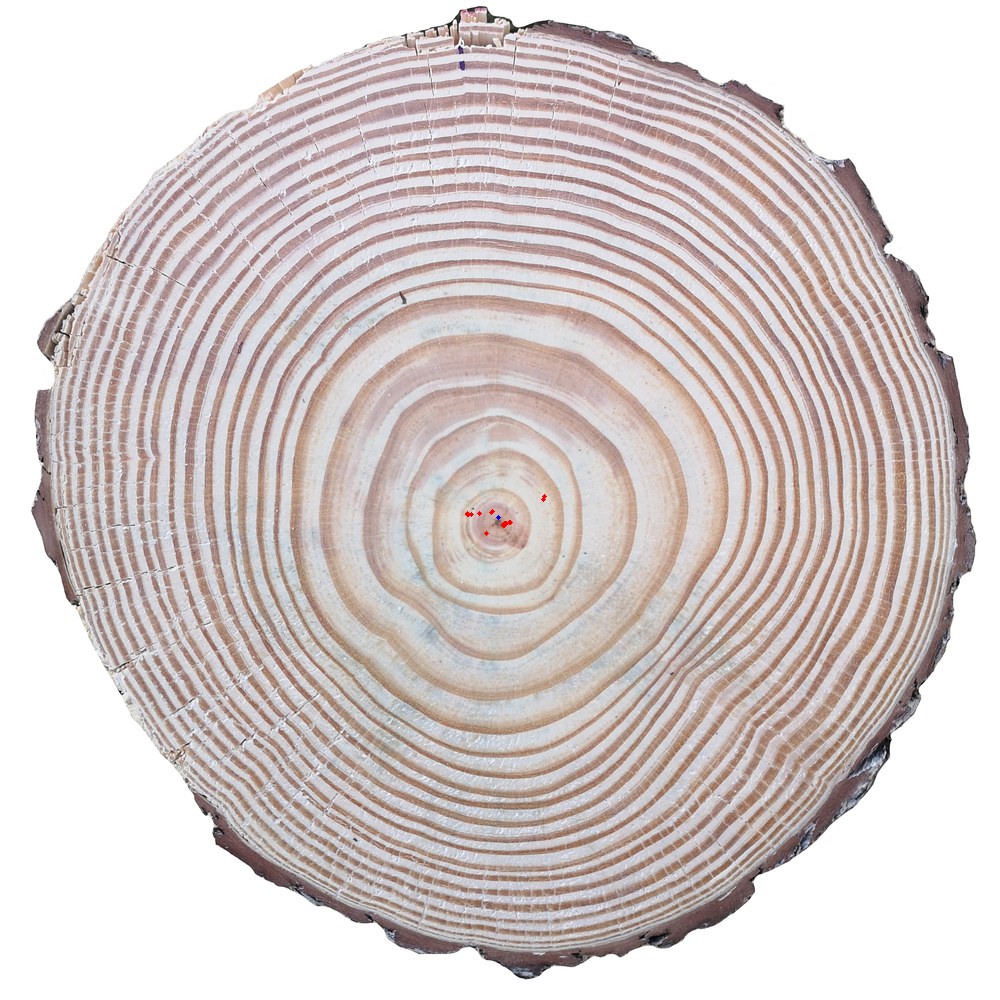}
            \caption{}
            \label{fig:peak}
        \end{subfigure}\\
        \vspace{10pt}
        \caption{Pith detector pipeline. (a) Masked input image. The background is colored in white. (b) Local Orientation output. The normals to the local patch orientation with good certainty score are drawn. (c) Accumulator Space. (d) Pith's prediction is colored in blue. Accumulator space maximums are colored in red.} 
         \label{fig:main_pipeline}
    \end{center} 
\end{figure}

Schraml et al. \cite{Schraml2013} proposed two algorithms for pith detection: the first is for RGB images, and the second for tomographic images. This article implements the first one, depicted in Algorithm \ref{algo:Globalalgo}. Figure \ref{fig:main_pipeline} illustrates the full pipeline. The method requires an RGB image of a wood slice and a background mask as input. Figure \ref{fig:img} shows the masked slice, i.e., without background. First, in the $local\_orientation\_estimation$ function, the image is divided into blocks, and for each one, a local estimation of the directions of the rings is made. The method also produces a certainty score that measures the confidence in estimating those local dominant orientations.  Figure \ref{fig:lo_lines} illustrates the normals to the local orientation estimation output. Only lines with certainty score higher than a parameter $lo\_certainty\_th$  are kept, i.e., the lines estimated with greater confidence. The closer $lo\_certainty\_th$ is to 1, the better. Then, in the $m\_accumulation\_space$ function, the normals to the dominant direction in each block are traced, and 
an accumulator space is defined to count the intersections of the normals. Figure \ref{fig:accumulator} shows the accumulator space. Finally, in the $find\_peak$ function, the position of the pith is the principal peak in the accumulation space. Figure \ref{fig:peak} illustrates the predicted peak in blue.  The accumulator space global maximums are colored in red.

\begin{algorithm}[!htbp]
  \KwIn{$Im_{in}$, // RGB wood cross-section image\\
  $mask$, // Background mask\\
  Parameters: \\
 $block\_overlap$, // Percentage of overlapping between patches. Patch Stage parameter\\
  $block\_width\_size$, // Defines the patch width. Patch Stage parameter\\
  $block\_height\_size$, // Defines the patch height. Patch Stage parameter\\
  $fft\_peak\_th$, // Frequency filter threshold ($\lambda$). Local orientation parameter\\
  $lo\_method$, // Local orientation method. Local orientation parameter\\
  $lo\_certainty\_th$, // Linear Symmetry threshold. Local orientation parameter\\ 
  $acc\_type$, // Lines accumulation method. Accumulator Space parameter\\
  $peak\_blur\_sigma$, // Gaussian blurring ($\sigma$). Peak parameter\\
   }
  \KwOut{Pixel pith location}
  \SetAlgoLined
  \LinesNumbered
  \BlankLine 
  $l\_{lo}$ $\leftarrow$ $local\_orientation\_estimation$($Im_{in}$, $mask$, $block\_overlap$, $block\_width\_size$, $block\_height\_size$, $fft\_peak\_th$, $lo\_method$, $lo\_certainty\_th$)\\
  $m\_accumulation\_space$ $\leftarrow$ $accumulation\_space$($l\_lo$, $acc\_type$)\\
  $peak$ $\leftarrow$ $find\_peak$($m\_accumulation\_space$, $peak\_blur\_sigma$)\\
     \KwRet{$peak$}
\caption{Pith Detection}
\label{algo:Globalalgo}
\end{algorithm}

\subsection{Local Orientation Estimation}
\begin{figure}
    \begin{center}
    \begin{subfigure}{0.3\textwidth}
        \includegraphics[width=\textwidth]{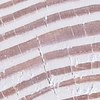}
        \caption{}
        \label{fig:block}
    \end{subfigure}
    \begin{subfigure}{0.3\textwidth}
        \includegraphics[width=\textwidth]{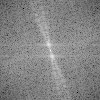}
        \caption{}
        \label{fig:fft_block}
    \end{subfigure}
    \begin{subfigure}{0.3\textwidth}
        \includegraphics[width=\textwidth]{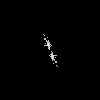}
        \caption{}
        \label{fig:fft_block_filtered}
    \end{subfigure}  
    \caption{(a) RGB image patch. (b) The Fourier Spectrum of the image patch shown in (a). (c) Preprocessed Fourier Spectrum of the same patch.} 
     \label{fig:lo_preprocessing}
    \end{center} 
\end{figure}

The principal hypothesis to locate the pith is that the normals to the local orientation of each ring cross the pith. To estimate the local orientation of the tree rings, the image is divided into blocks (patches). Inside each patch, the rings appear as roughly parallel curves, as shown in figure \ref{fig:block}.  To estimate their local orientation, the authors use the fact that these patches can be approximated to \textit{simple neighborhoods}, as defined in \cite{Granlund}: a 2D region in which the gray level changes along one direction and is roughly constant along the other. The Fourier transform of a perfect \textit{simple neighborhood} is a line with a similar orientation as the direction of maximal variation of the \textit{simple neighborhood}. As the image departs from a pure 2D sinusoid, this line spreads. The idea is to estimate the orientation of the line, which concentrates the energy in the Fourier transform of the patch. Figure \ref{fig:fft_block} shows the spectrum of the patch illustrated in \ref{fig:block}. Note how the energy in the Fourier domain is concentrated around a line formed by the two maximums and has the same orientation as the normals to the tree rings. The real image is not exactly a \textit{simple neighborhood}, as the rings are curves instead of lines, and produces a noisy structure in the Fourier transform space that is not exactly a line. However, the principal direction can still be estimated. To facilitate the detection of this direction, the Fourier transform is filtered, as shown in figure \ref{fig:fft_block_filtered}. Further details can be read in the reference paper.

\begin{figure}
    \begin{center}
        \begin{subfigure}{0.3\textwidth}
            \includegraphics[width=\textwidth]{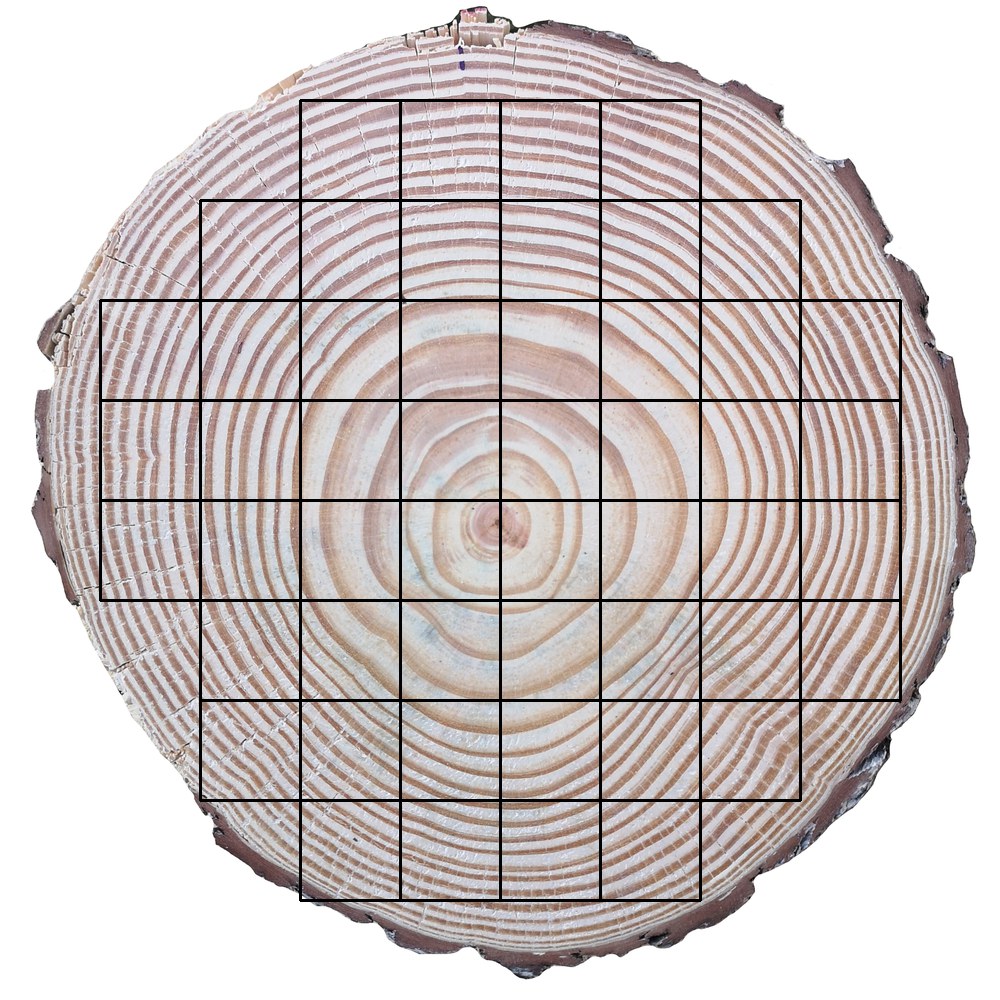}
            \caption{}
            \label{fig:patches}
        \end{subfigure}
        \vspace{10pt}
        \begin{subfigure}{0.3\textwidth}
            \includegraphics[width=\textwidth]{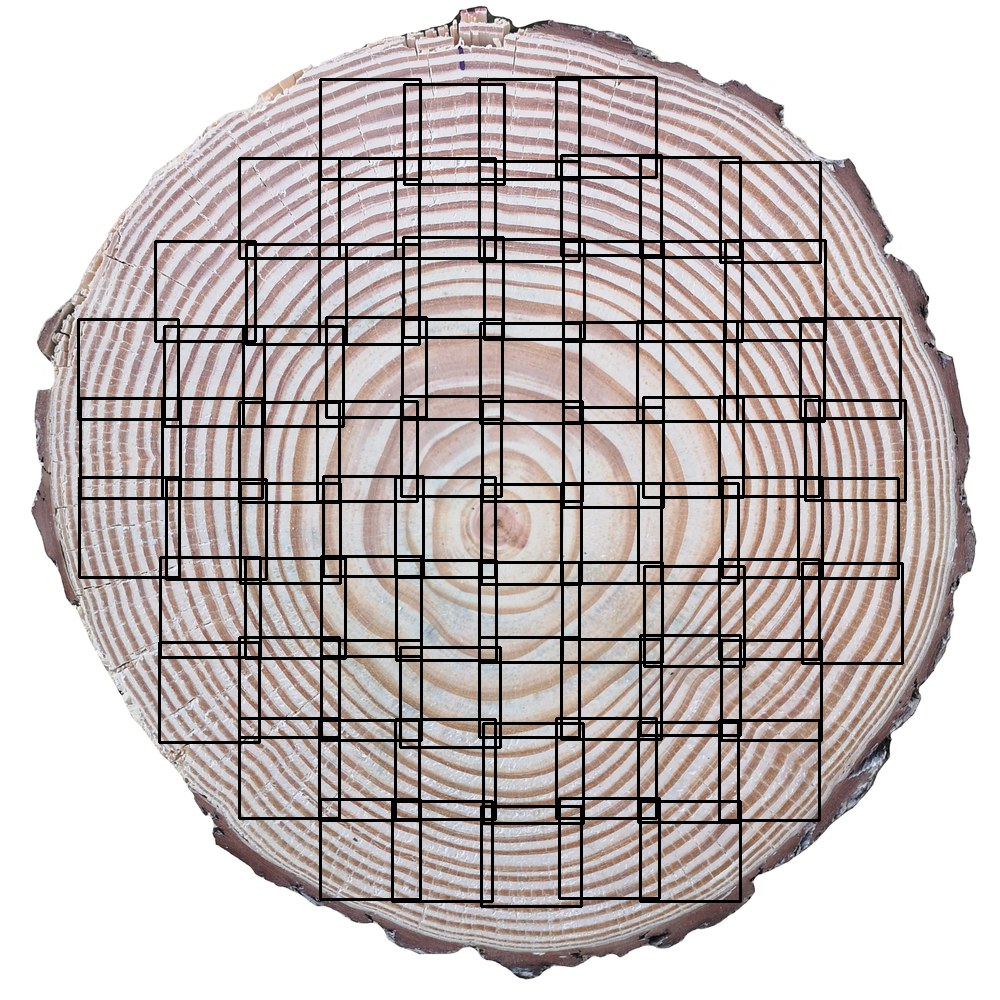}
            \caption{}
            \label{fig:patches_overlapping}
        \end{subfigure}
        \vspace{10pt}
        \begin{subfigure}{0.6\textwidth}
            \includegraphics[width=\textwidth]{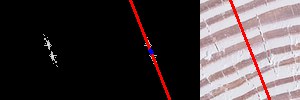}
            \caption{}
            \label{fig:fft_path_pca}
        \end{subfigure}\\
        \vspace{10pt}
        \begin{subfigure}{0.3\textwidth}
            \includegraphics[width=\textwidth]{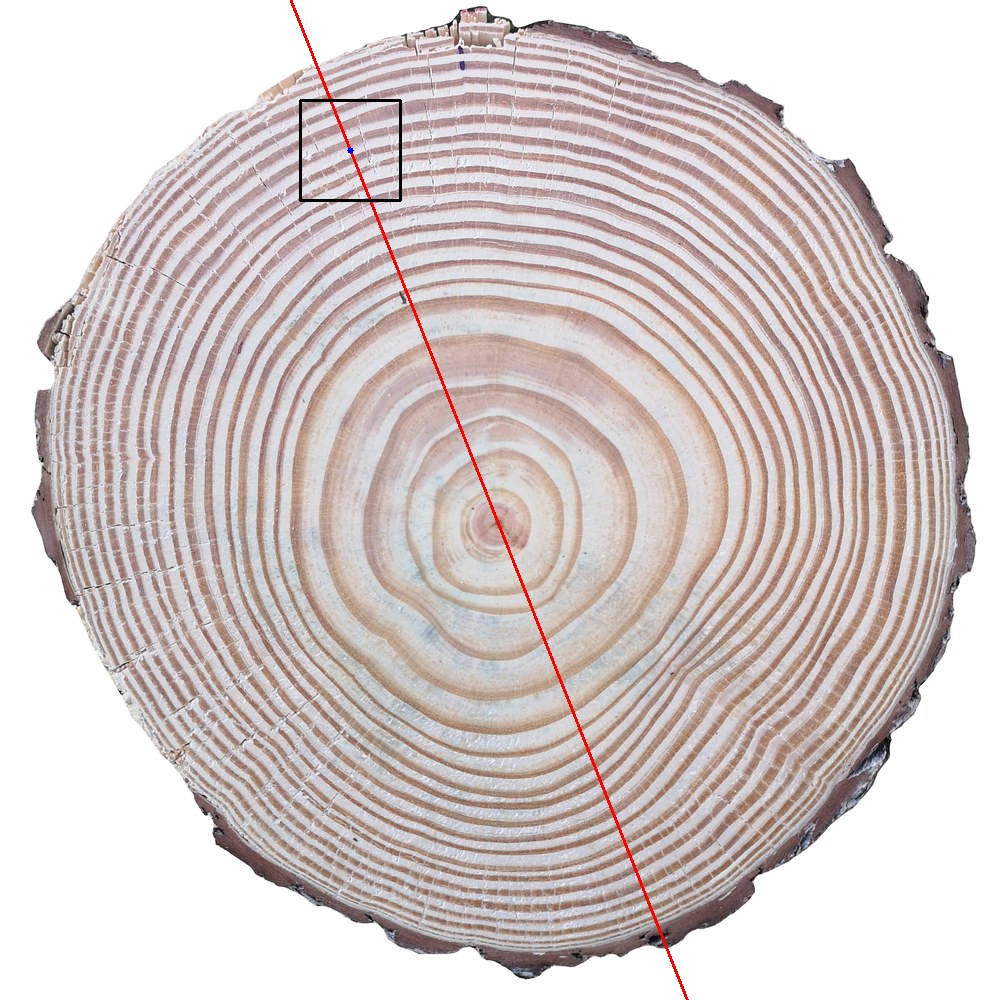}
            \caption{}
            \label{fig:fft_line}
        \end{subfigure}
        \vspace{10pt}
        \begin{subfigure}{0.3\textwidth}
            \includegraphics[width=\textwidth]{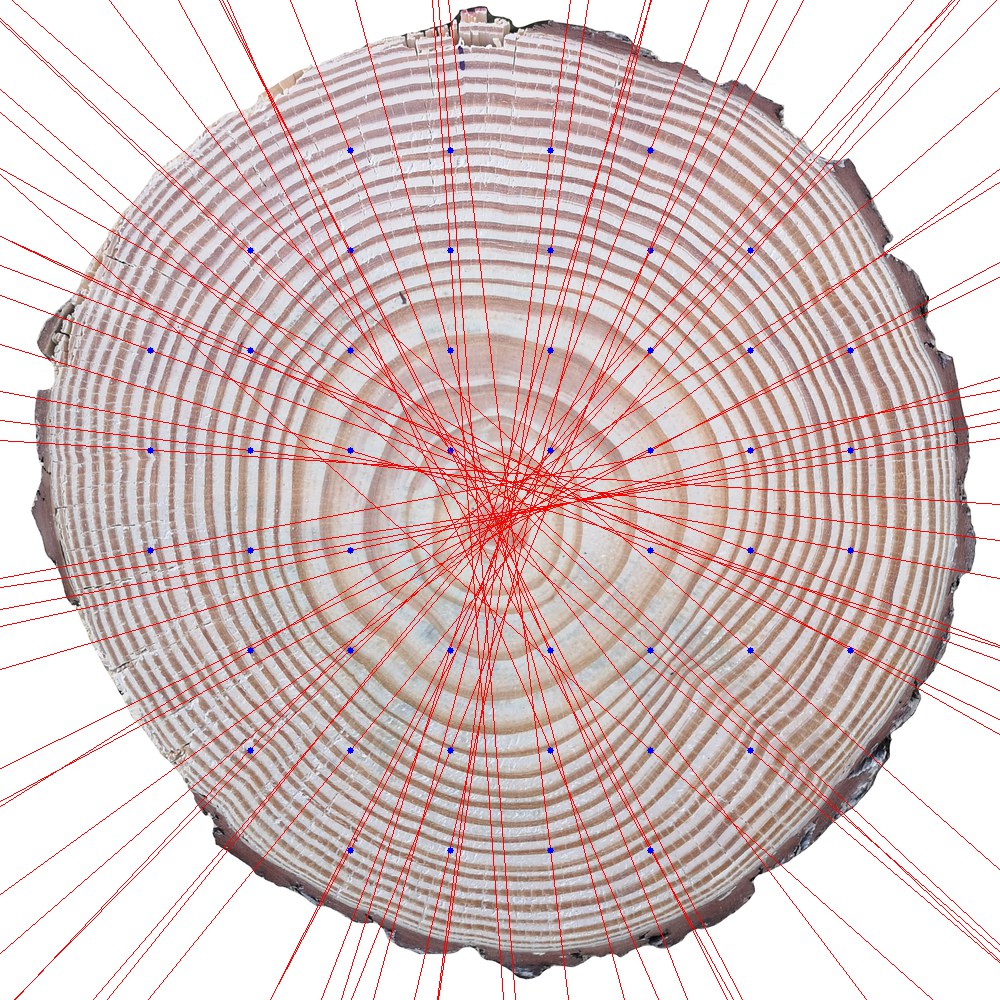}
            \caption{}
            \label{fig:all_lines}
        \end{subfigure}
        \vspace{10pt}
        \caption{Local orientation estimations steps. (a) The wood slice is split into non-overlapping patches. (b) Example of splitting wood cross-section image with $20\%$ overlapping patches. (c) Local estimation of the line. The left image shows the preprocessed 2D Fourier Transform of the patch. The center image shows the line, estimated by the PCA method, over the 2D Fourier Spectrum. The right image shows the estimated line superposed to the patch. (d) A line estimated within a given patch is traced over the whole slice. (e) The output of the local orientation estimation. Only the lines estimated with a certainty score higher than a threshold are traced. Blue dots indicate the patch centers. In this case, the pith location was estimated using non-overlapping patches.} 
   \label{fig:lo}
    \end{center} 
\end{figure}

Algorithm \ref{algo:lo} describes the local orientation detection method.  In line 1, the image is split into patches. Three parameters are used:
\begin{itemize}
    \item $block\_overlap$, define the overlapping between patches/blocks. It takes values between  0 and 1. Figure \ref{fig:patches} and Figure \ref{fig:patches_overlapping} illustrate patches with   $overlap=0\%$ and  $overlap=20\%$  respectively,
    \item $block\_width\_size$, define the block width size in pixels,
    \item $block\_height\_size$,  define the block height size in pixels.
\end{itemize}

\begin{algorithm}[!htbp]
  \KwIn{$Im_{in}$, // RGB wood cross-section image\\
  $mask$, // Background mask\\
  Parameters: \\
  $block\_overlap$, //Percentage of overlapping between patches. \\
  $block\_width\_size$, // Defines the patch width. \\
  $block\_height\_size$, // Defines the patch height. \\
  $fft\_peak\_th$, // Frequency filter threshold. \\
  $lo\_method$, // Local orientation method. \\
  $lo\_certainty\_th$, // Linear symmetry threshold. Formulation depends on $lo\_method$.\\
   }
  \KwOut{list of lines}
  \SetAlgoLined
  \LinesNumbered
  \BlankLine 
  $l\_blocks$, $l\_coordinates$ $\leftarrow$ $split\_image\_in\_blocks$($Im_{in}$, $mask$, $block\_overlap$, $block\_width\_size$, $block\_height\_size$)\\
  $l\_fs\_blocks$ $\leftarrow$ $compute\_fourier\_spectrum$($l\_blocks$)\\
  $l\_pre\_fs\_blocks$ $\leftarrow$ $preprocess\_fourier\_spectrum$($l\_fs\_blocks$, $fft\_peak\_th$ )\\
  $l\_lo$ $\leftarrow$ $lo\_linear\_symmetry$( $l\_pre\_fs\_blocks$, $lo\_method$)\\
  $l\_lo\_filtered$ $\leftarrow$ $filter\_lo\_by\_certainty$( $l\_lo$, $lo\_certainty\_th$)\\
  \KwRet{$l\_lo\_filtered$}
\caption{local\_orientation\_estimation}
\label{algo:lo}
\end{algorithm}

The Fourier transform of each patch is computed in line 2, and the Fourier spectrum of the patch is preprocessed in line 3. This stage includes two steps:
\begin{enumerate}
    \item A band-pass filter is applied with the low frequency defined as $\frac{block\_height\_size}{64}Hz$, and the high frequency defined as $\frac{block\_height\_size}{3}Hz$. These filter parameters are fixed once and for all. Here $block\_height\_size$ is the vertical dimension of the patch.
    \item The frequencies lower than $fft\_peak\_th$*\textbf{M}, where \textbf{M} is the maximum magnitude of the patch Fourier Spectrum, are filtered out.
\end{enumerate}

Figure \ref{fig:fft_block_filtered} illustrates the preprocessed Fourier Spectrum of a patch. In line 4, the local orientation is computed for each patch. Four methods are proposed for the line estimation, each one with a different way to estimate the certainty of the estimation:
\begin{itemize}
    \item $peak$: The line is defined as the one that connects the frequency with the highest magnitude in the Fourier space with the center. The certainty value is set to 1.
    \item $lsr$: local orientation is estimated by fitting a line, using a least squares method, for the values of the preprocessed patch Fourier Spectrum  (Figure \ref{fig:fft_block_filtered}). The certainty value is the absolute value of the coefficient of determination $|R^{2}|$.
    \item $wlsr$: local orientation is estimated fitting a line by a weighted least squares method for the values of the preprocessed patch Fourier Spectrum  (Figure \ref{fig:fft_block_filtered}). The weights are the square root of the Fourier magnitude. The certainty value is the absolute value of the coefficient of determination $|R^{2}|$.
    \item $pca$: Local orientation is estimated by extracting the orientation of the main direction using Principal Components Analysis. The certainty value is the ratio between the first and second eigenvalues. 
\end{itemize}

Finally, in line 5, we keep the local orientations with a certainty value greater than a $lo\_certainty\_th$ threshold. Figure \ref{fig:lo} illustrates the local orientation estimation steps.



\subsection{Accumulator Space}
The Accumulator Space (AS) is defined as a matrix of the same dimensions as the wood cross-section image. Two types of line accumulations are implemented: 
\begin{enumerate}
    \item Every pixel location where lines are intersected is incremented by one. Lines are represented in the plane by the equation a*x + b*y + c = 0.
    \item Every pixel location where a line passes through is increment by one. We used the function `line` from the OpenCV Python package to represent a line over an image. This function receives as parameter two pixel locations used as line extremes. In our case, we pass the intersection between the image axis and the line.
\end{enumerate}

Algorithm \ref{algo:acc} implements the accumulation space logic. As inputs, it gets the list of lines, $l\_lo$. It gets the parameter $type$ to indicate which type of line accumulator must be used. 

\begin{algorithm}[!htbp]
  \KwIn{$l\_lo$, // list of lines\\
  Parameter: \\
  $type$, // type of line accumulation.\\
   }
\KwOut{$m\_accumulation\_space$, // accumulator space matrix, same dimensions as the image}
  \SetAlgoLined
  \LinesNumbered
  \BlankLine 
 \If{$type$ $>$ $0$}{
   \tcc{Increment by one every pixel location where two lines are intersected}
    $m\_accumulation\_space$ $\leftarrow$ $lines\_intersection\_accumulation$($l\_lo$)
}
\Else{
    \tcc{Increment by one every pixel location where a line passes through}
    $m\_accumulation\_space$ $\leftarrow$ $lines\_pass\_through\_accumulation$($l\_lo$)
}
\KwRet{$m\_accumulation\_space$}
\caption{accumulation\_space}
\label{algo:acc}
\end{algorithm}

\subsection{Peak estimation}

A Gaussian filter of kernel size $\sigma$, is applied to the accumulator space before predicting the pith location. The predicted pith location corresponds to the global maximum. Their average location is returned if more than one global maxima is found.  

\section{Implementation}

\subsection{Installation}
Implementation is made in Python 3.11. Once the source code is downloaded, dependencies can be installed using the following command:

\begin{lstlisting}[language=bash]
  $ pip install -r requirements.txt
\end{lstlisting}

\subsection{Usage and parameters}
The program is executed using the following command:
\begin{lstlisting}[language=bash]
  $ python main.py --filename IMAGE_PATH  --output_dir OUTPUT_DIR
\end{lstlisting}

$IMAGE\_PATH$ refers to the directory where to search for the wood cross-section RGB image file, and $OUTPUT\_DIR$ to the path where the output file will be written.

The method has the following parameters, which can be set in the command line if needed
\begin{enumerate}
    \item --$new\_shape$ One-sided image length after resizing, in pixels. The resizing is always in a square format. The default value is 1000, meaning the resized image is $1000 \times 1000$ pixels.
    \item --$block\_width\_size$ width block size in pixels (default 100 pixels)
    \item --$block\_height\_size$ height block size in pixels (default 100 pixels)
    \item --$block\_overlap$ block overlapping (default 0.2, meaning 20\%)
    \item --$lo\_method$ Linear Orientation method (default PCA)
    \item --$lo\_certainty\_th$ Threshold to filter out less precise linear orientation (default 0.9)
    \item --$fft\_peak\_th$ Threshold ($\lambda$) to filter lower frequencies at the Fourier Spectrum stage (default 0.8)
    \item --$peak\_blur\_sigma$ Kernel blurring size ($\sigma$) applied over the accumulator space
    \item --$acc\_type$ accumulator line logic. If it is set to 1, line intersection logic is applied. If it is 0, every pixel where the line passes through is incremented by 1. 
\end{enumerate}

\section{Experiments and Results}
\subsection{Datasets}

\begin{figure}
\begin{centering}
    \begin{subfigure}{0.3\textwidth}
    \begin{centering}
    \includegraphics[width=\textwidth]{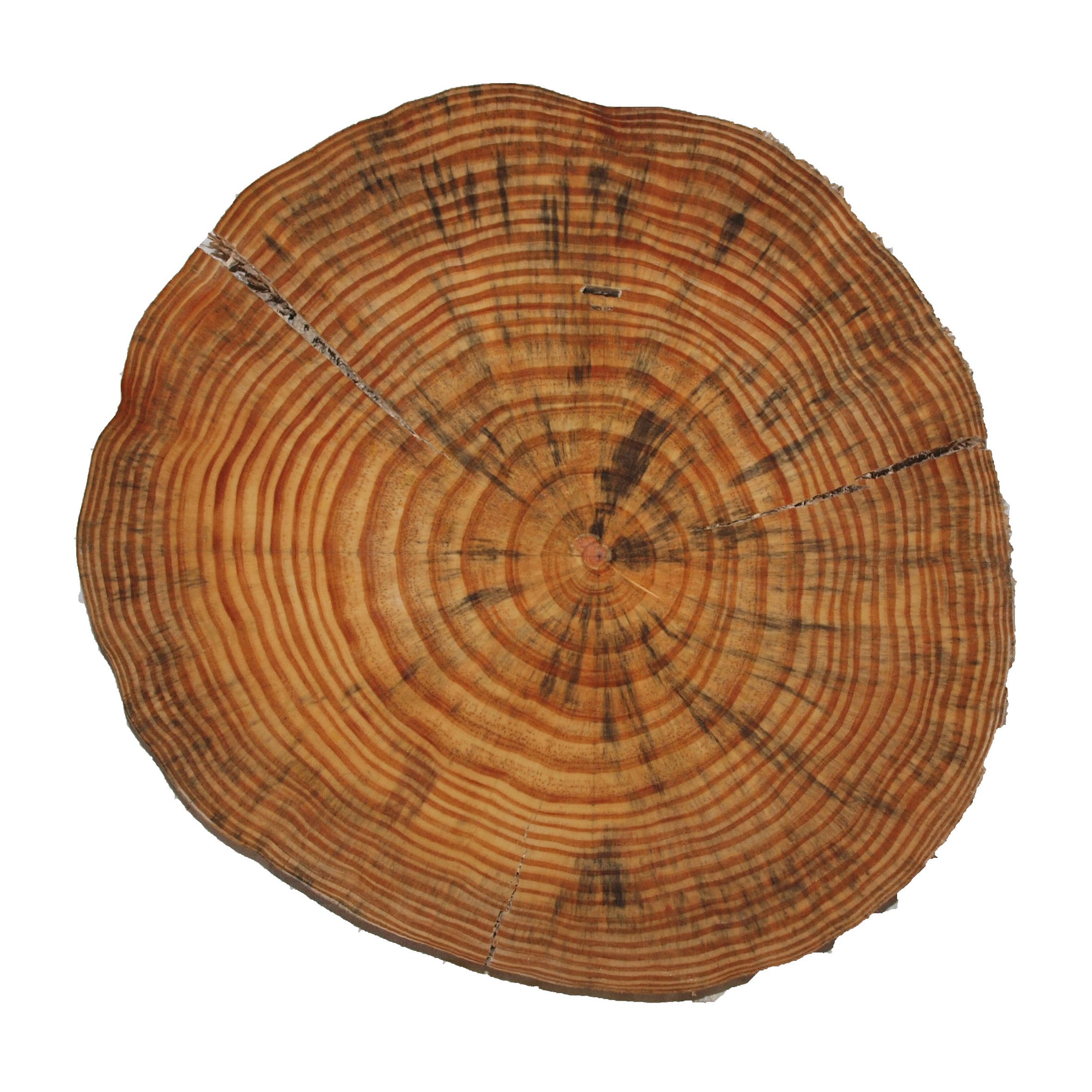}
    \label{fig:ddbb-F02a}
    \caption{F02a}
    \end{centering}
    \end{subfigure}
    \hfill
    \begin{subfigure}{0.3\textwidth}
    \begin{centering}
    \includegraphics[width=\textwidth]{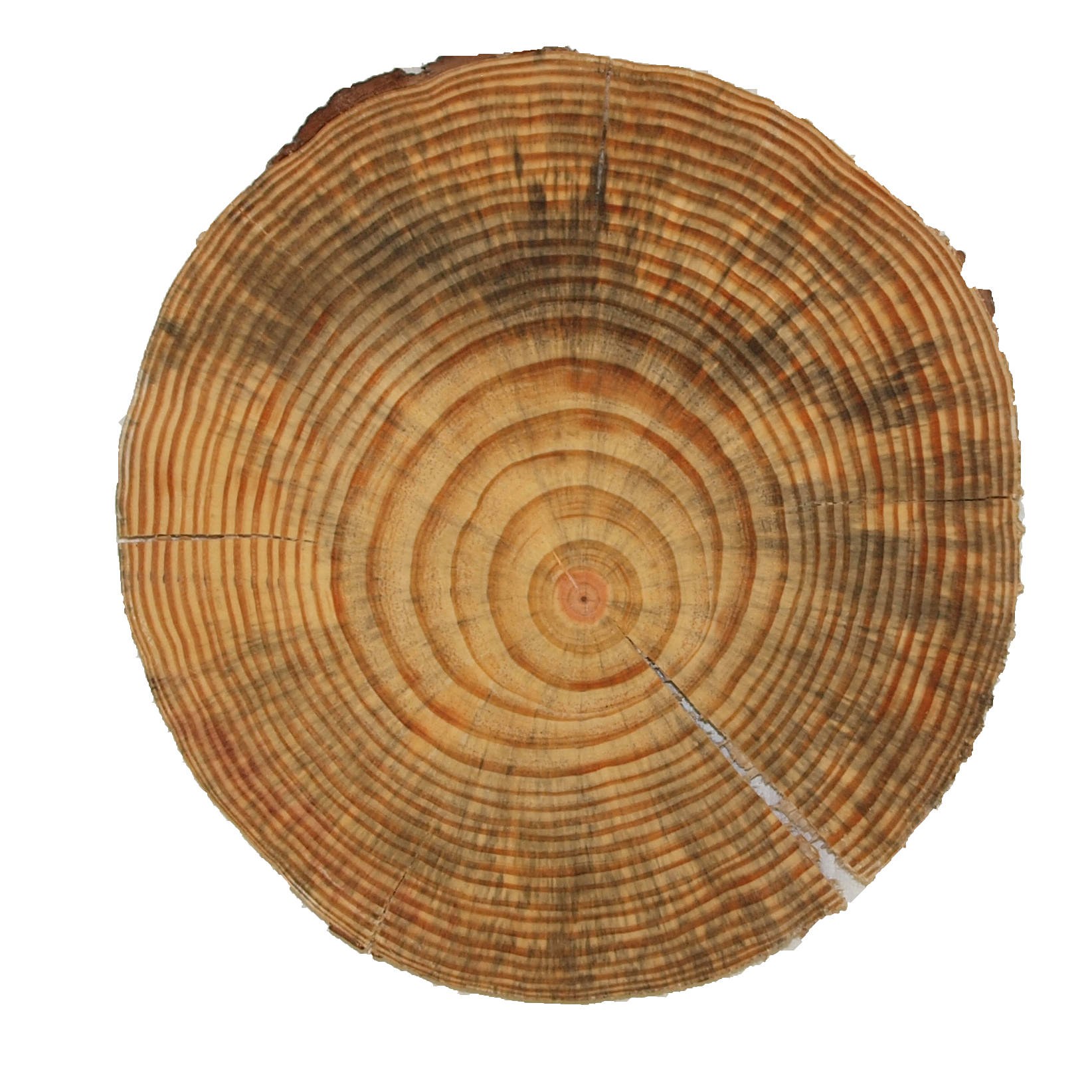}
    \label{fig:ddbb-F02b}
    \caption{F02b}
    \end{centering}
    \end{subfigure}
    \hfill
    \begin{subfigure}{0.3\textwidth}
    \begin{centering}
    \includegraphics[width=\textwidth] {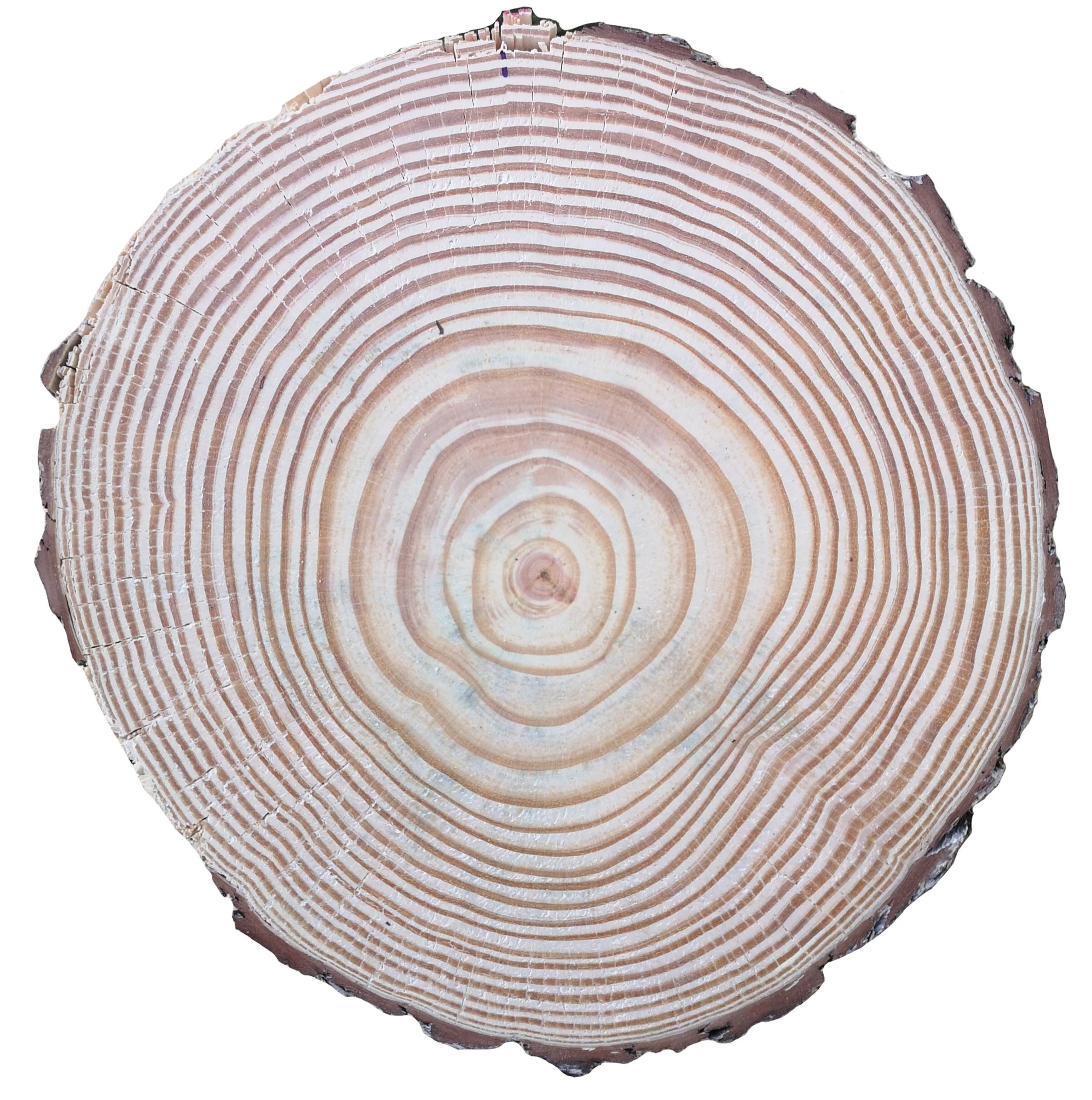}
    \label{fig:ddbb-F02c}
    \caption{F02c}
    \end{centering}
    \end{subfigure}
    \hfill

    \begin{subfigure}{0.3\textwidth}
    \begin{centering}
   \includegraphics[width=\textwidth]{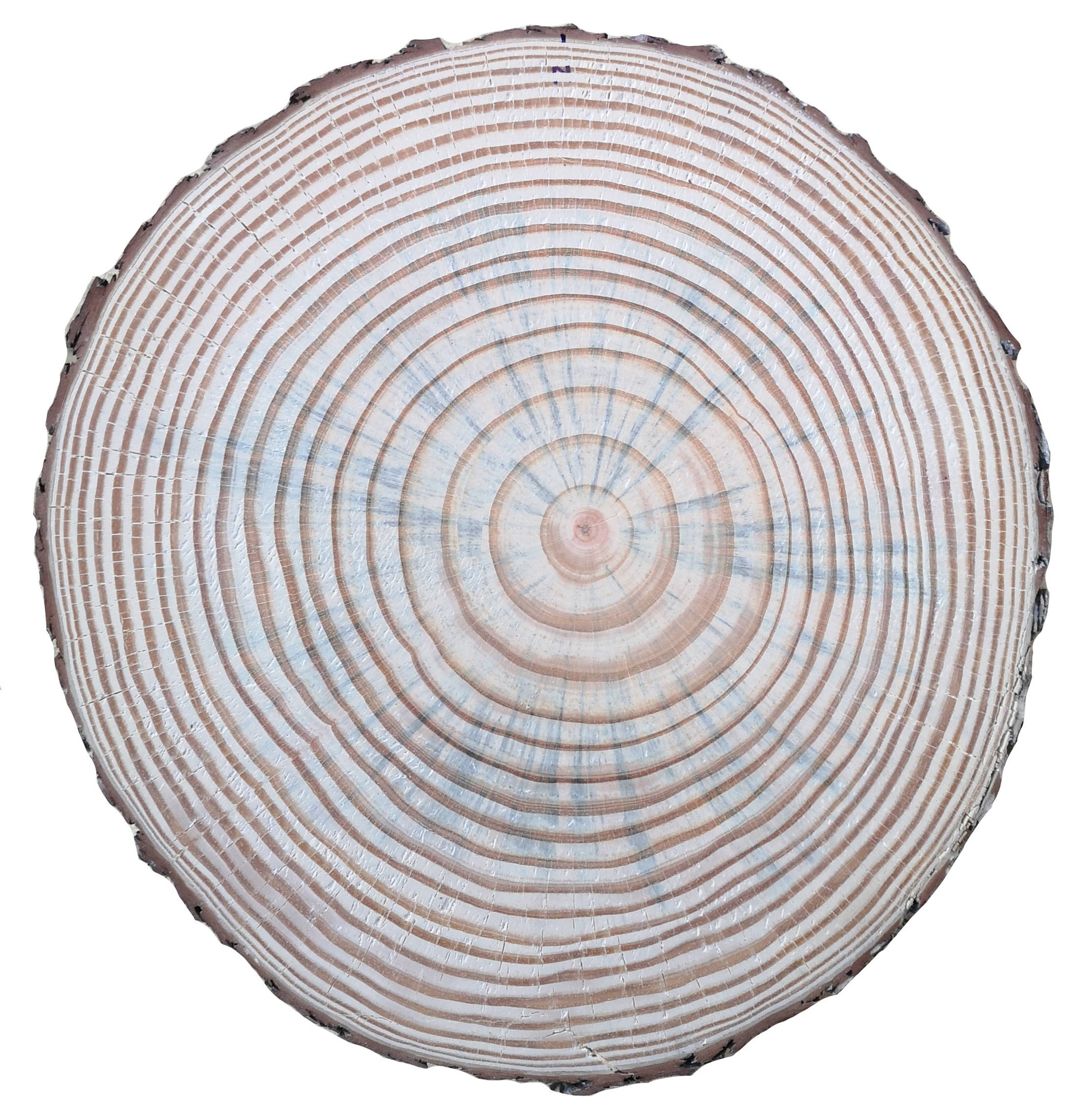}
    \label{fig:ddbb-F02d}
    \caption{F02d}
    \end{centering}
    \end{subfigure}
    \hfill
    \begin{subfigure}{0.3\textwidth}
    \begin{centering}
   \includegraphics[width=\textwidth]{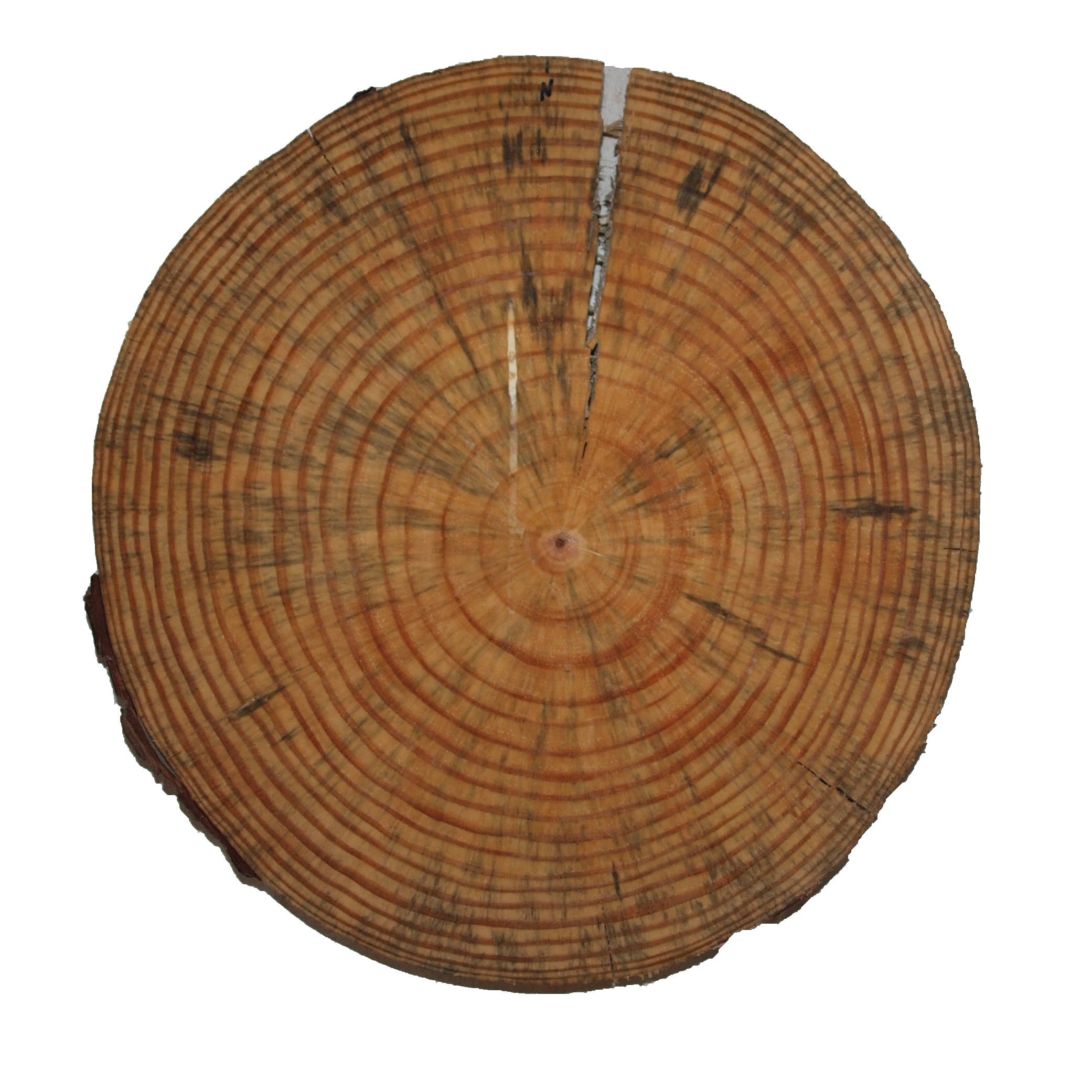}
    \label{fig:ddbb-F02e}
    \caption{F02e}
    \end{centering}
    \end{subfigure}
    \hfill
    \begin{subfigure}{0.3\textwidth}
    \begin{centering}
   \includegraphics[width=\textwidth]{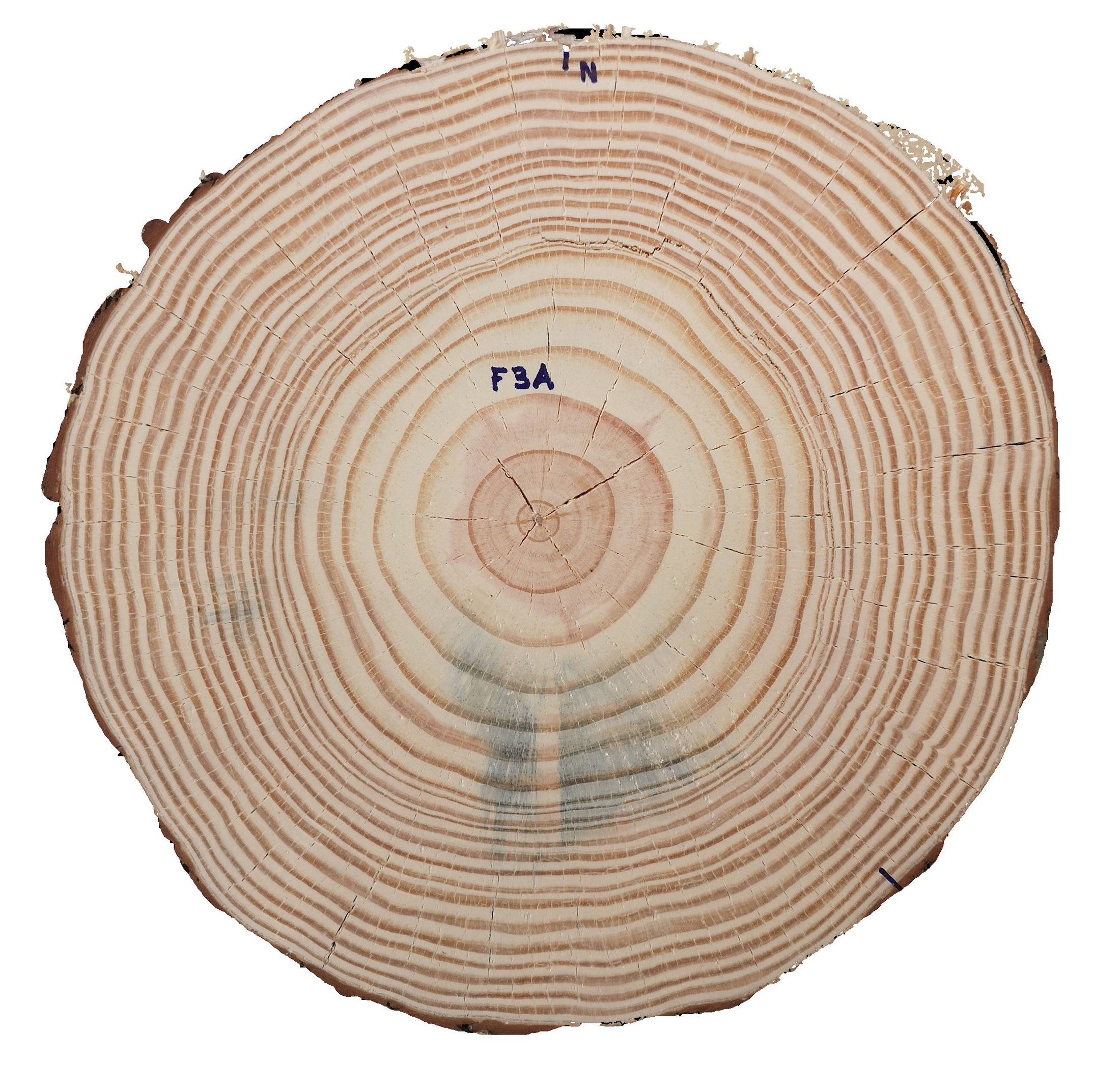}
    \label{fig:ddbb-F03c}
    \caption{F03c}
    \end{centering}
    \end{subfigure}
    \hfill
   
    \begin{subfigure}{0.3\textwidth}
    \begin{centering}
   \includegraphics[width=\textwidth]{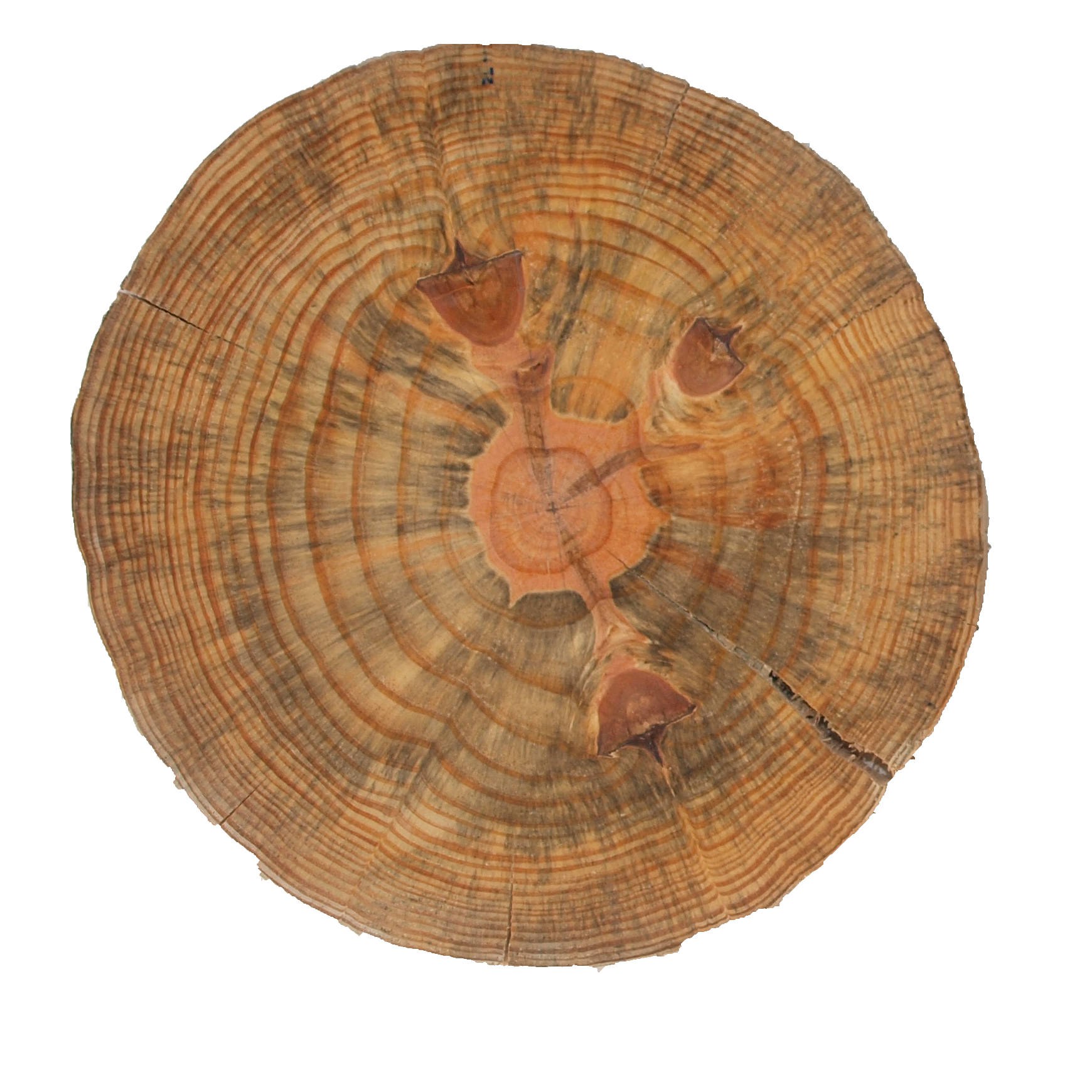}
    \label{fig:ddbb-F07b}
    \caption{F07b}
    \end{centering}
    \end{subfigure}
    \hfill
    \begin{subfigure}{0.3\textwidth}
    \begin{centering}
   \includegraphics[width=\textwidth]{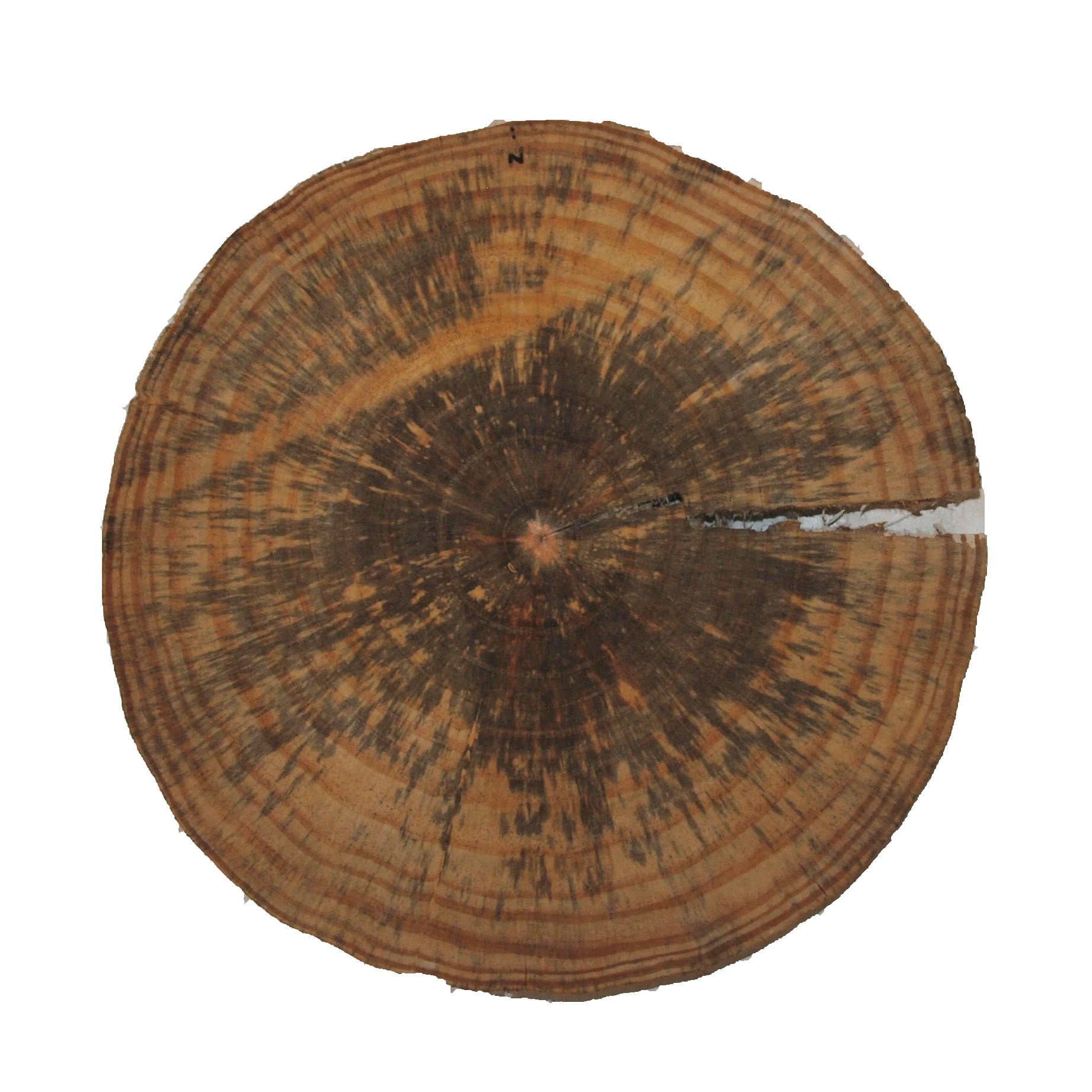}
    \label{fig:ddbb-L02b}
    \caption{L02b}
    \end{centering}
    \end{subfigure}
    \hfill
    \begin{subfigure}{0.3\textwidth}
    \begin{centering}
   \includegraphics[width=\textwidth]{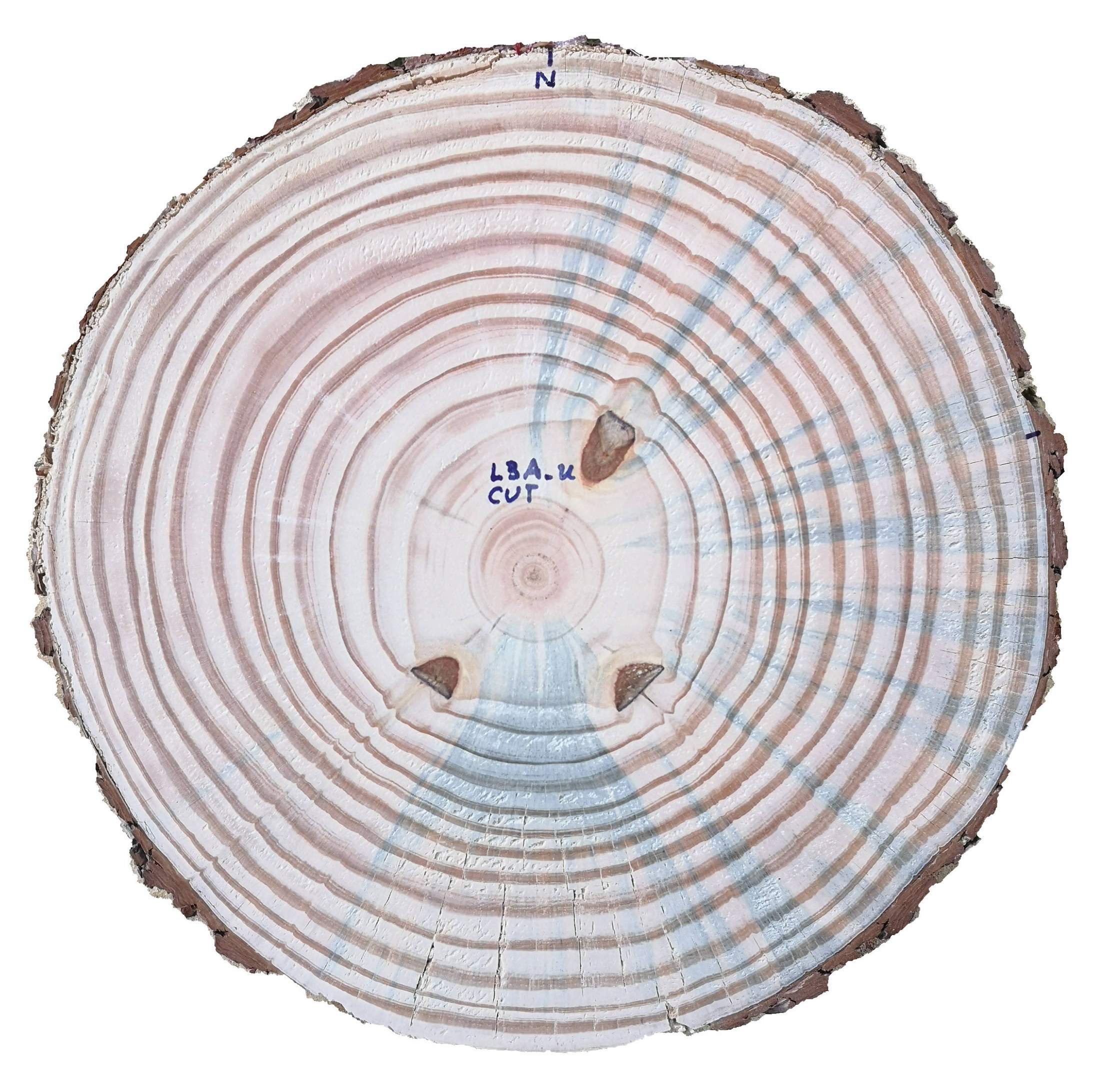}
    \label{fig:ddbb-L03c}
    \caption{L03c}
    \end{centering}
    \end{subfigure}
    \hfill
   \caption{Some examples of the images in the UruDendro dataset. Note the variability of the images and the presence of fungus (for example, in image L02b), knots (for example, in images F07b and F03c), and cracks (for example, in images F02e and L03c). }
   \label{fig:ddbb}
\end{centering}
\end{figure}

\begin{figure}
\begin{centering}
    \begin{subfigure}{0.3\textwidth}
    \begin{centering}
    \includegraphics[width=\textwidth]{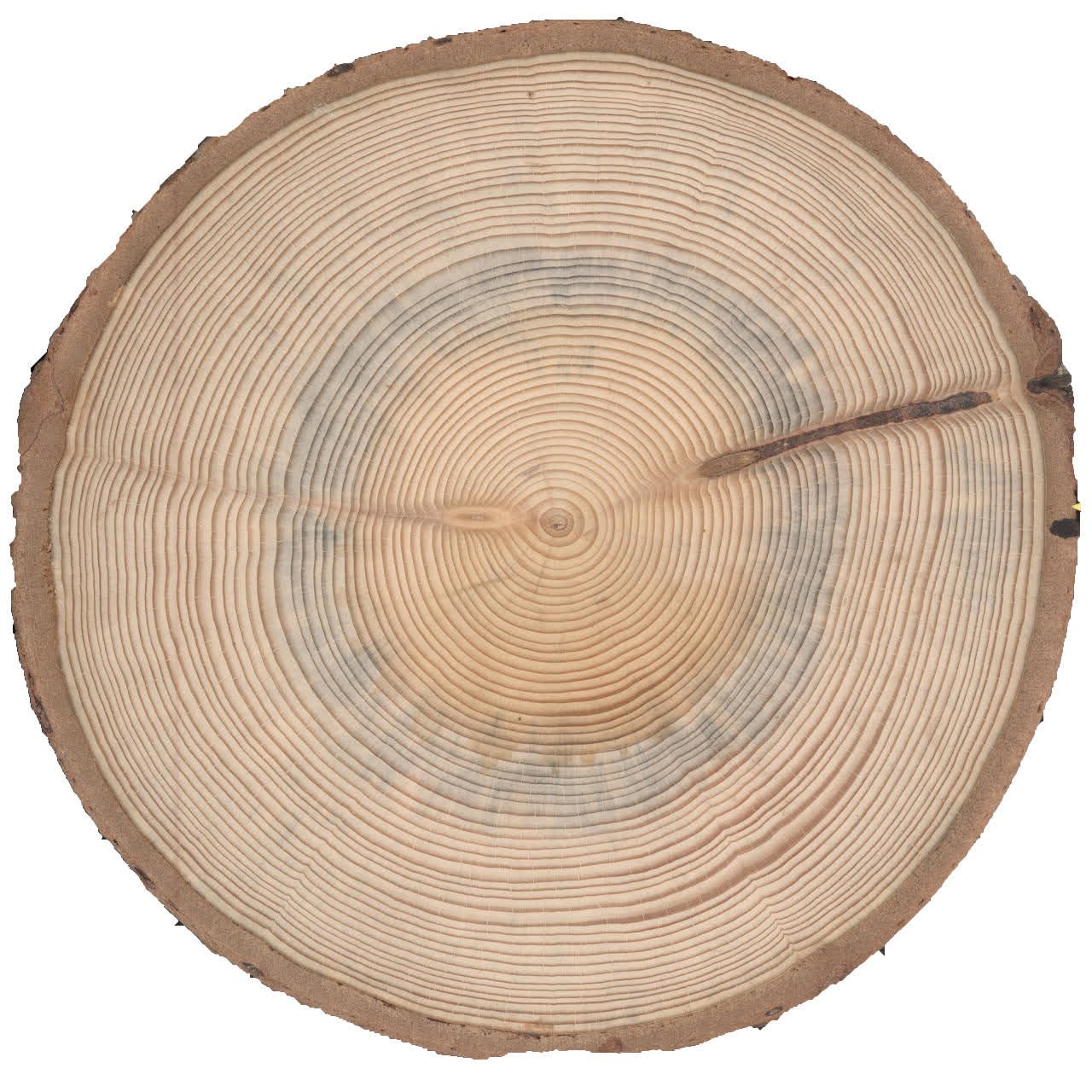}
    \label{fig:ddbb-ac1}
    \caption{AbiesAlba1}
    \end{centering}
    \end{subfigure}
    \hfill
    \begin{subfigure}{0.3\textwidth}
    \begin{centering}
    \includegraphics[width=\textwidth]{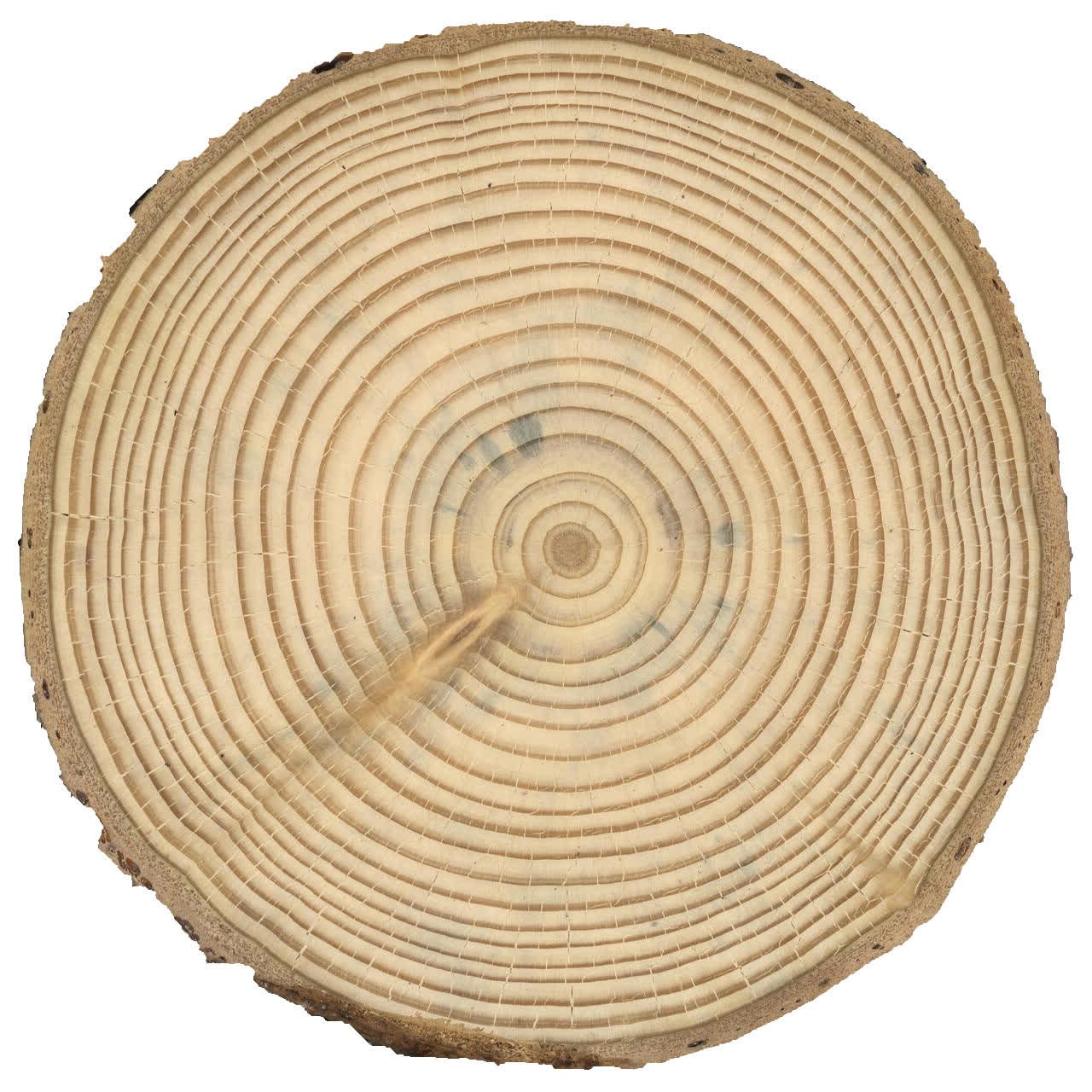}
    \label{fig:ddbb-ac2}
    \caption{AbiesAlba2}
    \end{centering}
    \end{subfigure}
    \hfill
    \begin{subfigure}{0.3\textwidth}
    \begin{centering}
    \includegraphics[width=\textwidth]{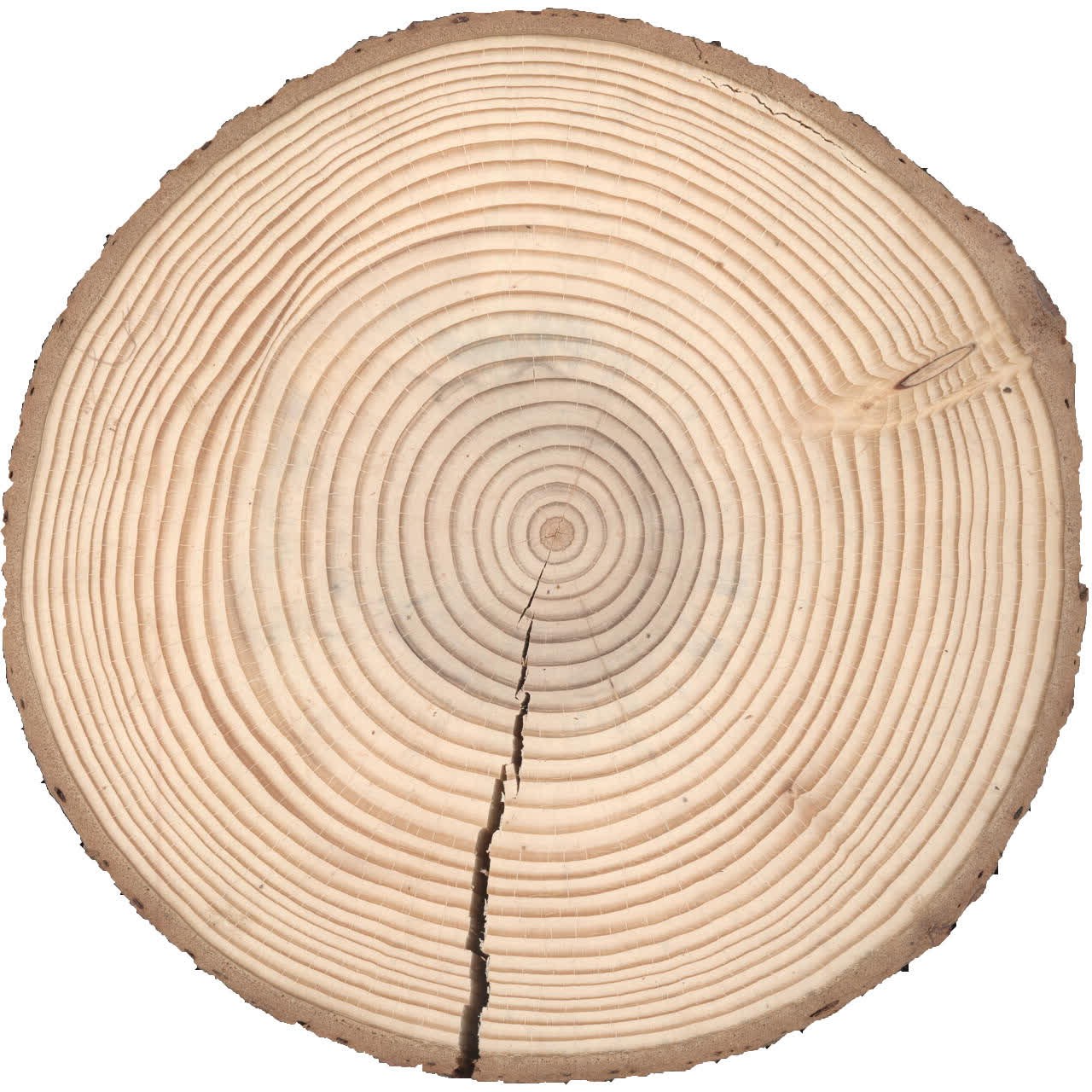}
    \label{fig:ddbb-ac3}
    \caption{AbiesAlba3}
    \end{centering}
    \end{subfigure}
    \hfill
    \begin{subfigure}{0.3\textwidth}
    \begin{centering}
   \includegraphics[width=\textwidth]{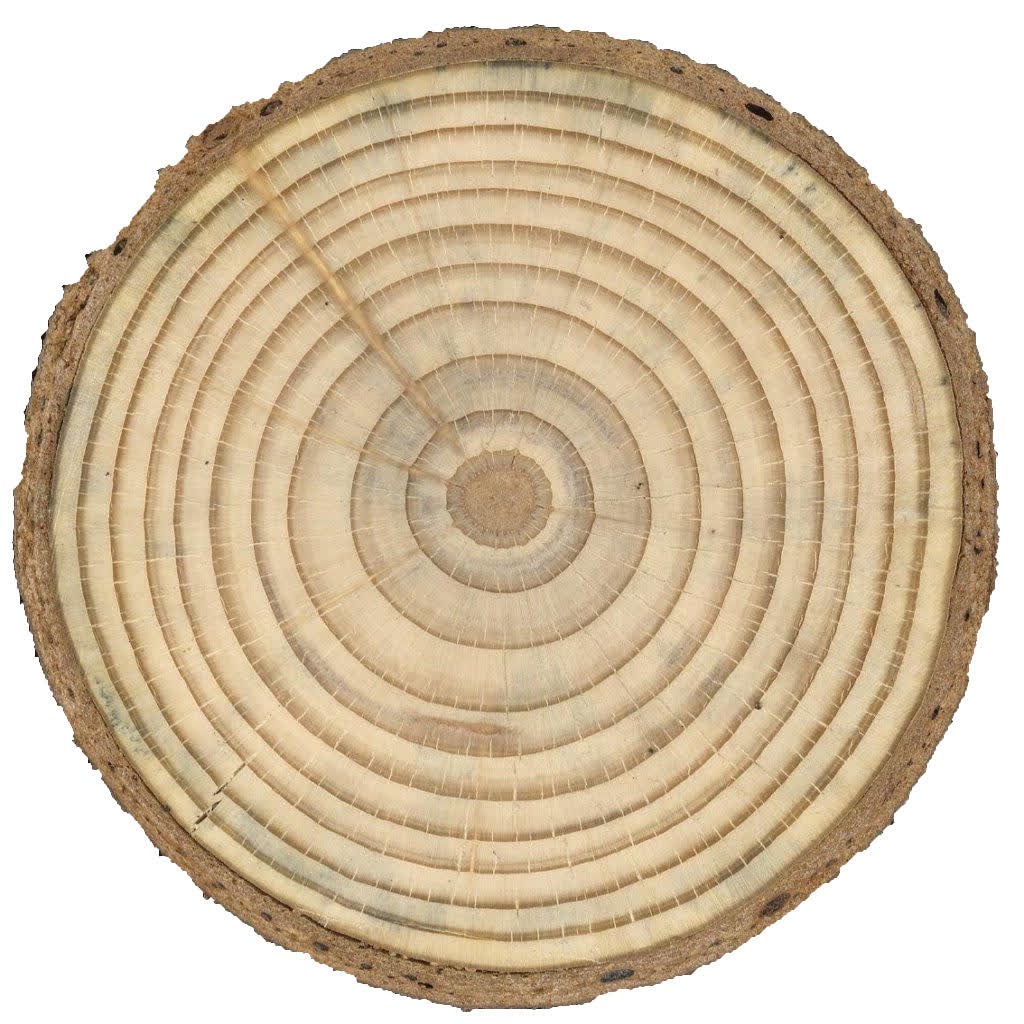}
    \label{fig:ddbb-ac4}
    \caption{AbiesAlba4}
    \end{centering}
    \end{subfigure}
    \hfill
    \begin{subfigure}{0.3\textwidth}
    \begin{centering}
   \includegraphics[width=\textwidth]{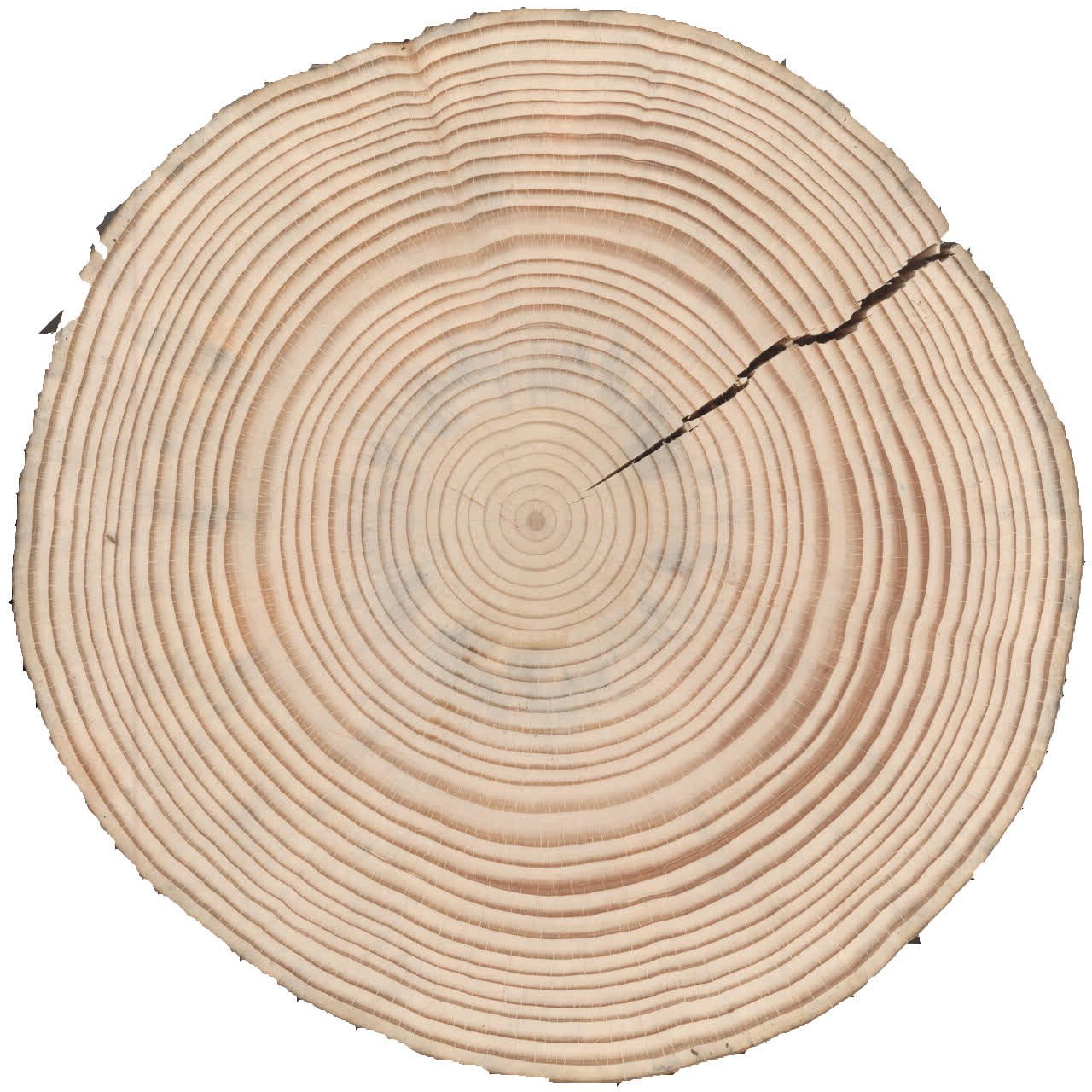}
    \label{fig:ddbb-ac5}
    \caption{AbiesAlba5}
    \end{centering}
    \end{subfigure}
    \hfill
    \begin{subfigure}{0.3\textwidth}
    \begin{centering}
   \includegraphics[width=\textwidth]{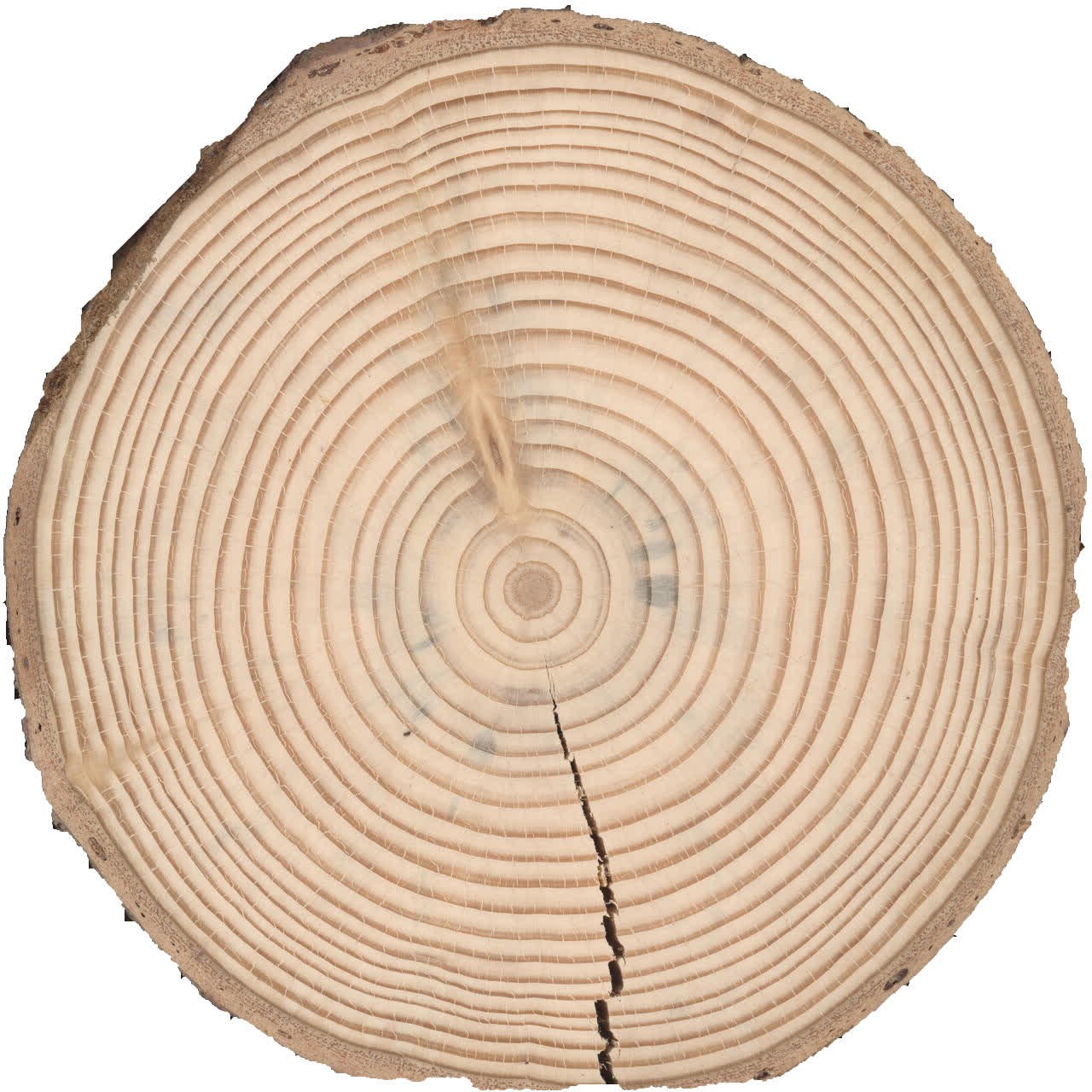}
    \label{fig:ddbb-ac6}
    \caption{AbiesAlba6}
    \end{centering}
    \end{subfigure}
    \hfill
    \begin{subfigure}{0.3\textwidth}
    \begin{centering}
   \includegraphics[width=\textwidth]{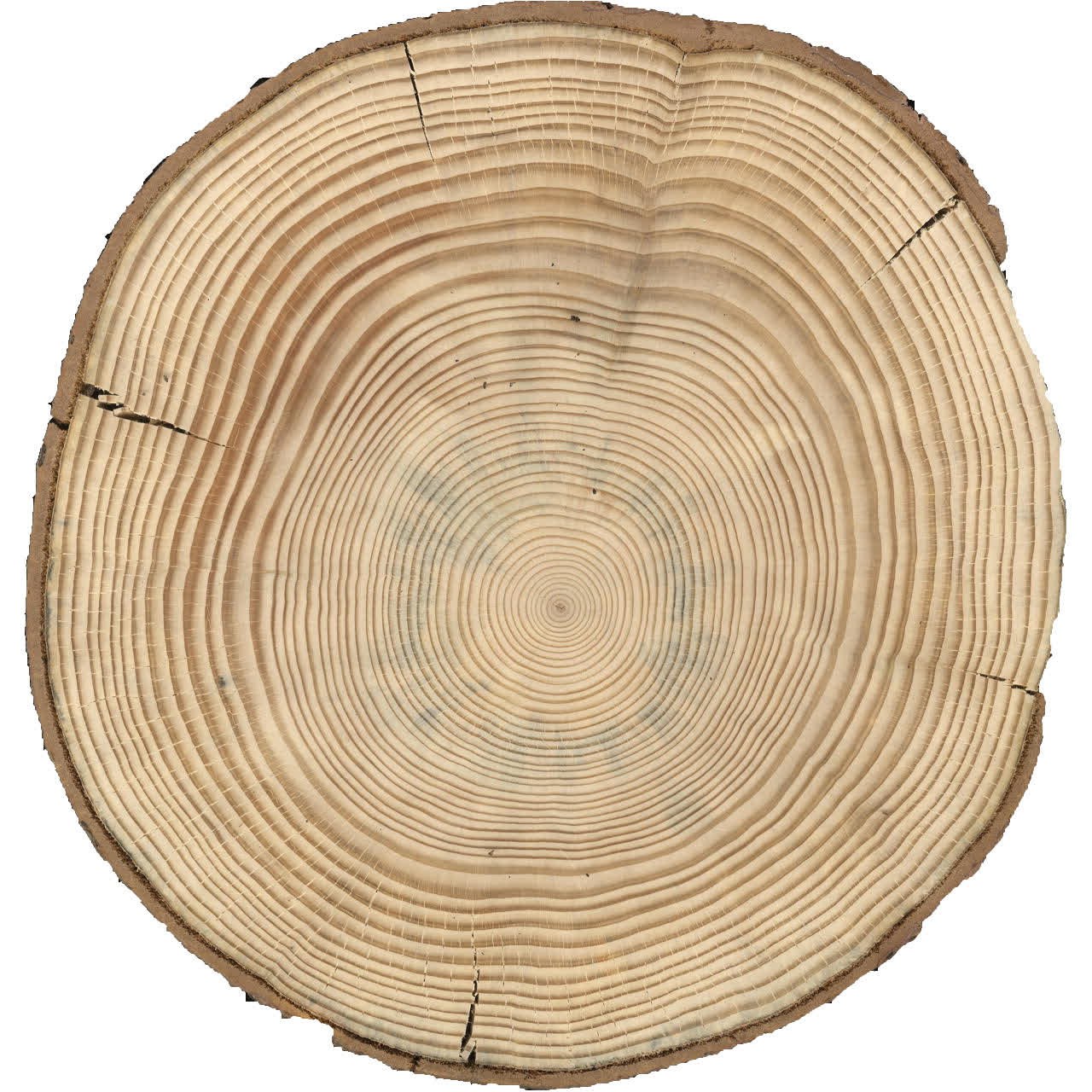}
    \label{fig:ddbb-ac7}
    \caption{AbiesAlba7}
    \end{centering}
    \end{subfigure}

   \caption{Images from Kennel at al. \cite{KennelBS15} dataset.}
   \label{fig:ddbb_ac}
\end{centering}
\end{figure}

To test the method, we use the following datasets:
\begin{enumerate}
    \item  \textbf{UruDendro} dataset. An online public database \cite{UruDendro} with images of cross-sections of commercially grown \textit{Pinus taeda} trees from northern Uruguay, ranging from 13 to 24 years old, composed of fourteen individual trees collected in February 2020 in Uruguay. Slices were about 5 to 20 cm thick and were dried at room temperature without further preparation.  Due to the drying process, radial cracks and blue fungus stains in the wood slices appear. Surfaces were smoothed with a handheld planer and a rotary sander. Photographs were taken under different lighting conditions; some were photographed indoors and moistened to maximize contrast between early- and late-wood. Other pictures were taken outdoors.  
    The dataset has 64 images of different resolutions, from $913 \times 900$ to $2877 \times 2736$ pixels. The collection contains several challenging features for automatic ring detection, including illumination and surface preparation variation, fungal infection (blue stains), knot formation, missing bark and interruptions in outer rings, and radial cracking.
    Figure \ref{fig:ddbb} shows some images in this UruDendro dataset. Tree ring delineation and pith location are available (annotated manually by experts).
    \item \textbf{Kennel} dataset. Kennel et al. \cite{KennelBS15} made available a public dataset of 7 images of Abies alba and presented a method for detecting tree rings. Figure \ref{fig:ddbb_ac} illustrates the dataset. Tree ring delineation and pith location are available (annotated manually by experts).
\end{enumerate}

\subsection{Metrics}
The Euclidean distance between prediction and ground truth is used to evaluate the method.
\begin{equation}
    Dist(dt, gt) = \sqrt{(dt_x - gt_x)^{2} + (dt_y - gt_y)^{2}}
    \label{equ:euclidean}
\end{equation}

Where $dt=(dt_x, dt_y)$ refers to the pith prediction and $gt=(gt_x, gt_y)$ refers to the pith ground truth. A normalized version of the former metric is defined below 

\begin{equation}
    Norm\_dist(dt, gt) = 1000 \frac{Dist(dt, gt)}{d}
    \label{equ:normalized}
\end{equation}

Where $d$ represents the maximum diameter of the disk passing through the center, the factor $1000$ is arbitrarily chosen. To compute $d$, 180 lines crossing through the ground truth pith are traced. They are equally separated between them by 2 degrees. The length of the longest one defines $d$.

In both UruDendro and Kennel datasets, the rings are annotated. This allows us to use the following metric. If the pith detection is within the pith region (inside the trunk medulla), the detection is considered a True Positive (TP). In contrast, the detection is assigned as False Positive (FP) if it is outside. Then, the Precision metric is computed as:

\begin{equation}
    Precision = \frac{TP}{TP+FP}
    \label{equ:presicion}
\end{equation}

\subsection{Experiments}
We compare the method with the one proposed by Decelle et al. \cite{Decelle2022} (called ACO from here on), as it achieves state-of-the-art results and the code is available.  We used the default  parameters of that method.

For the implemented Schraml and Uhl method (called Schraml from here on), we use the parameter exploration described in Table \ref{tab:method_grid_search}. We optimize the method performance over the following parameters: patch size, patch overlapping percentage, local orientation method, and local orientation threshold. The other parameters are fixed. Additionally, Table \ref{tab:method_grid_search} shows the parameter configuration that performs the best over each dataset using the minimum Euclidean distance to the ground truth, Equation \ref{equ:euclidean}.

\begin{table}[]
\centering
\begin{tabular}{|c|c|c|c|}
\hline
Parameter          & Values for grid search & UruDendro & Kennel\\ \hline
Patch size (pixels)    & 20, 30, 50, 100     & 50  & 25 \\ \hline
Overlapping (\%)       & 0, 20, 40, 50      &   50 & 50 \\ \hline
Local orientation method          & PCA, Peak     &Peak    & PCA \\ \hline
Local orientation threshold      & 0.75, 0.85, 0.95 &  0.75 &  0.75 \\ \hline
Image Size (pixels)    & 1000x1000        & 1000x1000 & 1000x1000\\ \hline
$\lambda$   ($fft\_peak\_th$)       & 0.6            & 0.6  & 0.6 \\ \hline
Accumulator Method & Addition & Addition      & Addition \\ \hline
Gaussian filter $\sigma$           & 3  & 3    &  3          \\ \hline
\end{tabular}
\caption{Parameters used for the Schraml method. In the second column are the parameters used for the exploration. Only square patches are used. The parameter configuration in the third and fourth columns gives the best results for the named dataset.}
\label{tab:method_grid_search}
\end{table}

\subsection{Results}

Table \ref{tab:urudendro_results_d1} shows the results over the UruDendro dataset for the Euclidean  (Equation \ref{equ:euclidean}) and the normalized distances (Equation \ref{equ:normalized}). The Schraml method performs better than the ACO method in Mean, 90 percentile (P90), 95 percentile  (P95), and Max statistics. Table \ref{tab:kennel_results_d1} shows the results over the Kennel dataset. Over this dataset, the ACO method performs slightly better. 


\begin{table}[]
\centering
\begin{tabular}{|l|cc|cc|}
\hline
                 & \multicolumn{2}{c|}{\textbf{Distance}}                                        & \multicolumn{2}{c|}{\textbf{Normalized distance}}                                        \\ \hline
\textbf{Statics} & \multicolumn{1}{l|}{\textbf{ACO}} & \multicolumn{1}{l|}{\textbf{Schraml}} & \multicolumn{1}{l|}{\textbf{ACO}} & \multicolumn{1}{l|}{\textbf{Schraml}} \\ \hline
\textbf{Mean}    & \multicolumn{1}{c|}{72}        & 44                                & \multicolumn{1}{c|}{46}        & 26                                \\ \hline
\textbf{Std}     & \multicolumn{1}{c|}{192}       & 29                                 & \multicolumn{1}{c|}{131}       & 17                                 \\ \hline
\textbf{Median}  & \multicolumn{1}{c|}{10}        & 39                                 & \multicolumn{1}{c|}{6}         & 22                                 \\ \hline
\textbf{P90}     & \multicolumn{1}{c|}{226}       & 76                                 & \multicolumn{1}{c|}{115}       & 48                                 \\ \hline
\textbf{P95}     & \multicolumn{1}{c|}{307}       & 94                                 & \multicolumn{1}{c|}{176}       & 50                                 \\ \hline
\textbf{Max}     & \multicolumn{1}{c|}{1024}      & 143                                & \multicolumn{1}{c|}{690}       & 95                                 \\ \hline
\end{tabular}
\caption{Result over the UruDendro dataset using the Euclidean distance between the position of the detected and GT pith (Equation \ref{equ:euclidean}) in the 2nd and 3rd columns, and the Normalized Euclidean distance (Equation \ref{equ:normalized}) in the last two columns.}
\label{tab:urudendro_results_d1}
\end{table}


\begin{table}[]
\centering
\begin{tabular}{|l|cc|cc|}
\hline
                 & \multicolumn{2}{c|}{\textbf{Distance}}                                        & \multicolumn{2}{c|}{\textbf{Normalized distance}}                                        \\ \hline
\textbf{Statics} & \multicolumn{1}{l|}{\textbf{ACO}} & \multicolumn{1}{l|}{\textbf{Schraml}} & \multicolumn{1}{l|}{\textbf{ACO}} & \multicolumn{1}{l|}{\textbf{Schraml}} \\ \hline
\textbf{Mean}    & \multicolumn{1}{c|}{6}            & 8                                     & \multicolumn{1}{c|}{7}            & 7                                     \\ \hline
\textbf{Std}     & \multicolumn{1}{c|}{2}            & 7                                     & \multicolumn{1}{c|}{4}            & 6                                     \\ \hline
\textbf{Median}  & \multicolumn{1}{c|}{6}            & 8                                     & \multicolumn{1}{c|}{6}            & 6                                     \\ \hline
\textbf{P90}     & \multicolumn{1}{c|}{8}            & 17                                    & \multicolumn{1}{c|}{10}           & 15                                    \\ \hline
\textbf{P95}     & \multicolumn{1}{c|}{8}            & 17                                    & \multicolumn{1}{c|}{12}           & 16                                    \\ \hline
\textbf{Max}     & \multicolumn{1}{c|}{9}            & 17                                    & \multicolumn{1}{c|}{14}           & 18                                    \\ \hline
\end{tabular}
\caption{Results over the Kennel dataset using the Euclidean distance between the position of the detected and GT pith (Equation \ref{equ:euclidean}) in columns 2 and 3. Normalized euclidean distance (Equation \ref{equ:normalized}) results are shown in the last 2 columns.}
\label{tab:kennel_results_d1}
\end{table}

The former metric shows how close the detection is to the pith ground truth in pixels. However, in some applications, it is enough if the detection is within the pith region (the medulla). The precision metric (equation \ref{equ:presicion}) gives an idea of this situation. Table \ref{tab:kennel_results_p} shows the results over the Kennel dataset, where both methods perform similarly. On the other hand, over the UruDendro dataset, the ACO method performs much better than Schraml, with a precision of 70\% vs 14\%. 

\begin{figure}
\begin{centering}
    \begin{subfigure}{0.4\textwidth}
    \begin{centering}
    \includegraphics[width=\textwidth]{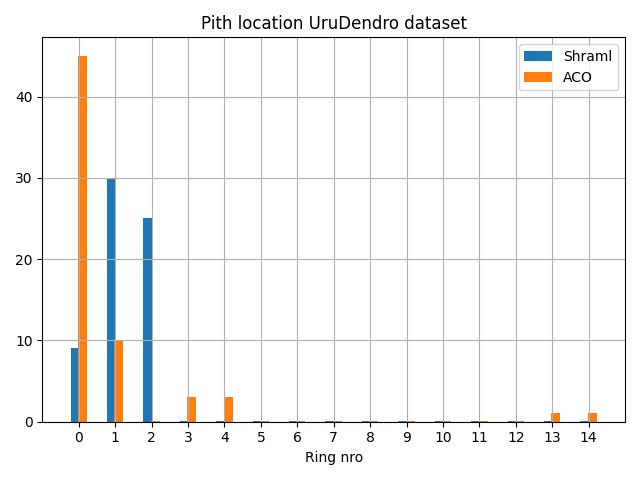}
    \label{fig:density_urudendro}
    \caption{}
    \end{centering}
    \end{subfigure}
    \hfill
    \begin{subfigure}{0.4\textwidth}
    \begin{centering}
    \includegraphics[width=\textwidth]{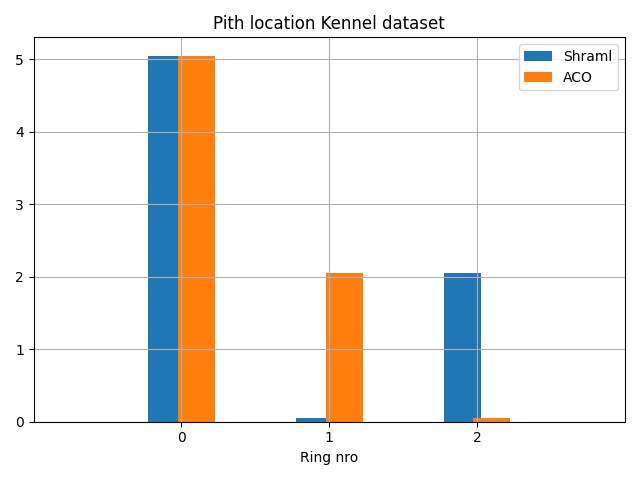}
    \label{fig:density_kennel}
    \caption{}
    \end{centering}
    \end{subfigure}
   \caption{Ring region where each detected pith falls. A histogram plot is generated with UruDendro (a) and Kennel (b)  results.}
   \label{fig:density}
\end{centering}
\end{figure}

Figure \ref{fig:density} shows a histogram from the ring region where the detected pith falls. Figure \ref{fig:density}.a refers to the UruDendro dataset. Over this dataset, ACO seems to be more precise (more than 40 detections fall inside the ring 0 region) but with a larger tail than the Schraml method. All the detections for the Schraml method fall between ring 0 (the medulla) and ring 2.
On the other hand, in the ACO method, the spread is much larger, with some detections located in rings 13 and 14.  Figure \ref{fig:density}.b illustrates the detected pith location over the Kennel dataset; for this, both methods perform very well.

\begin{table}[]
\centering
\begin{tabular}{|c|c|c|}
\hline
               & ACO & Schraml \\ \hline
TP             & 5   & 5      \\ \hline
FP             & 2   & 2      \\ \hline
Precision (\%) & 71  & 71     \\ \hline
\end{tabular}
\caption{Result over Kennel dataset using the precision metric (Equation \ref{equ:presicion}). }
\label{tab:kennel_results_p}
\end{table}

\begin{table}[]
\centering 
\begin{tabular}{|c|c|c|}
\hline
               & ACO & Schraml \\ \hline
TP             & 45  & 19     \\ \hline
FP             & 9   & 55     \\ \hline
Precision (\%) & 70  & 14     \\ \hline
\end{tabular}
\caption{Result over the UruDendro dataset using the precision metric (Equation \ref{equ:presicion}). }
\label{tab:urudendro_results_p}
\end{table}


\begin{figure}
\begin{centering}
    \begin{subfigure}{0.3\textwidth}
    \begin{centering}
    \includegraphics[width=\textwidth]{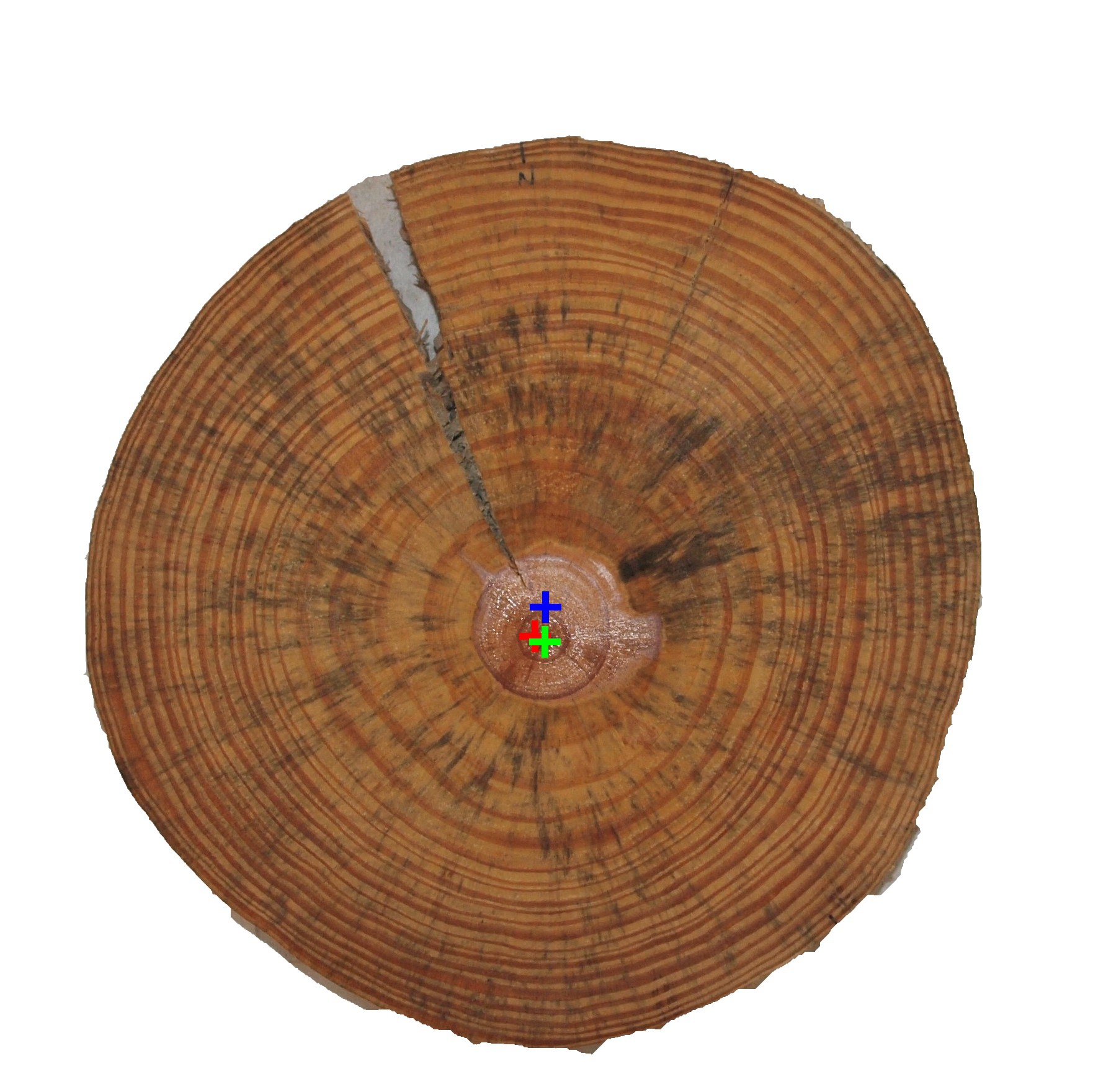}
    \label{fig:F04b_comp}
    \caption{F04b}
    \end{centering}
    \end{subfigure}
    \hfill
    \begin{subfigure}{0.3\textwidth}
    \begin{centering}
    \includegraphics[width=\textwidth]{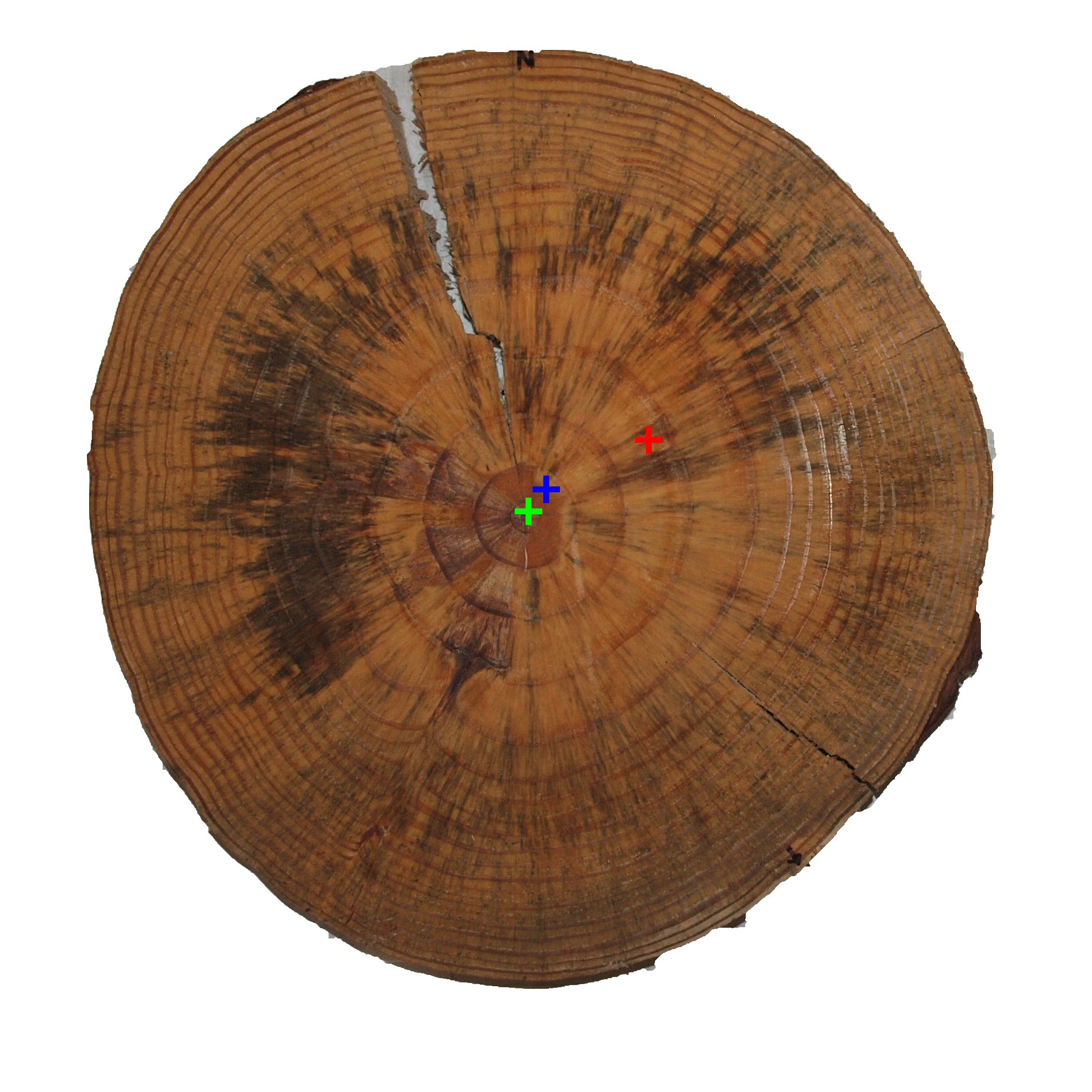}
    \label{fig:F07e_comp}
    \caption{F07e}
    \end{centering}
    \end{subfigure}
    \hfill
    \begin{subfigure}{0.3\textwidth}
    \begin{centering}
    \includegraphics[width=\textwidth]{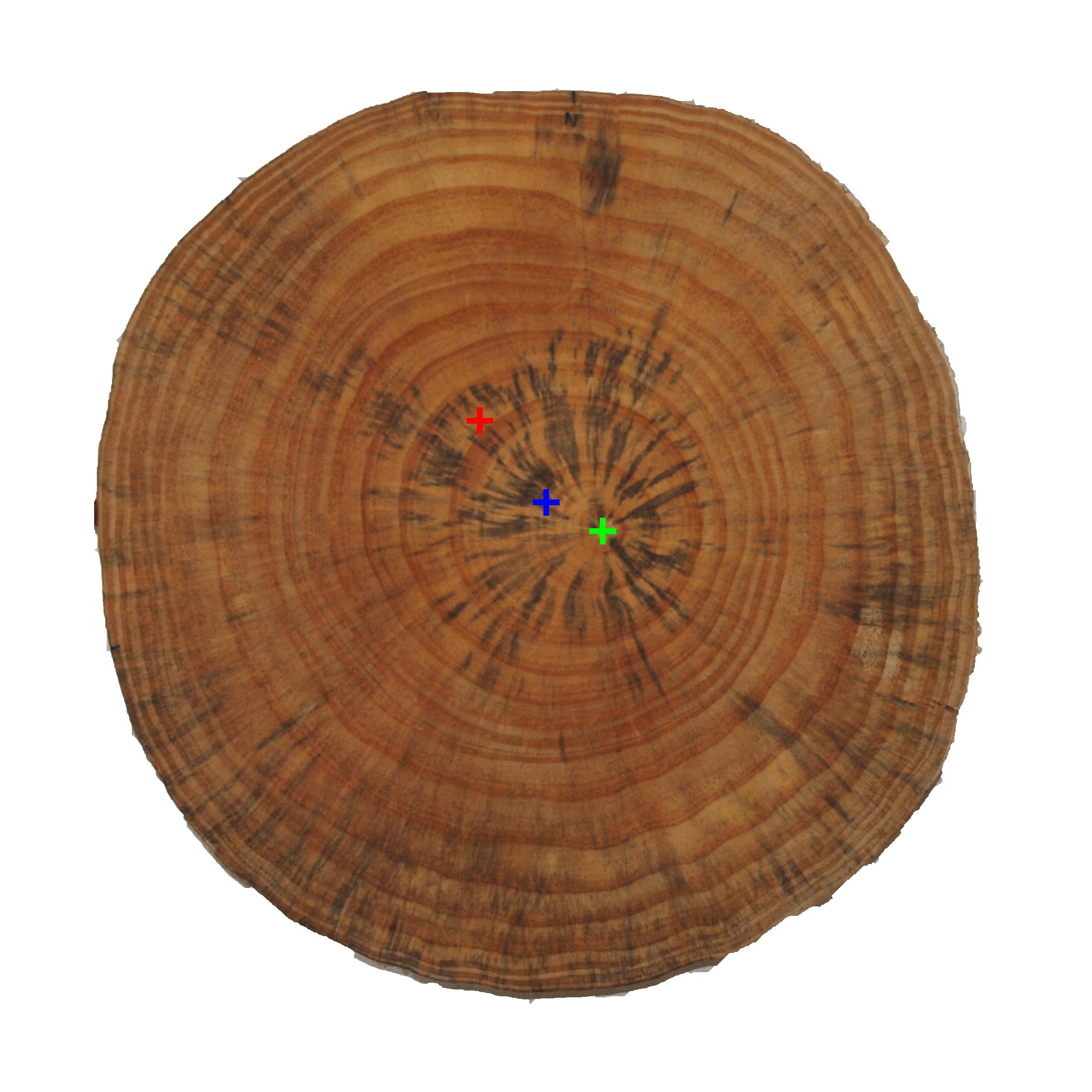}
    \label{fig:L02a_comp}
    \caption{L02a}
    \end{centering}
    \end{subfigure}
    \hfill
    \begin{subfigure}{0.3\textwidth}
    \begin{centering}
   \includegraphics[width=\textwidth]{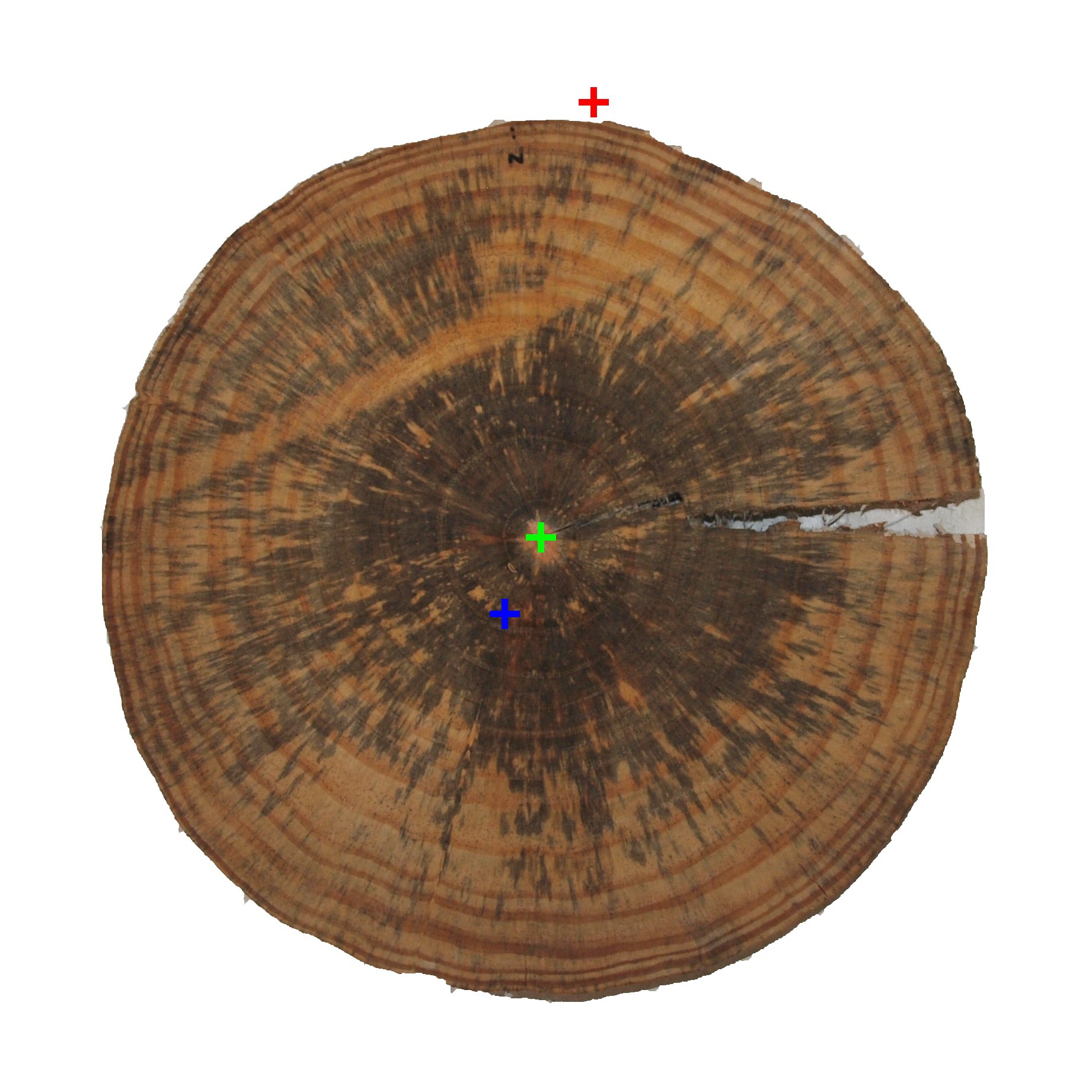}
    \label{fig:L02b_comp}
    \caption{L02b}
    \end{centering}
    \end{subfigure}
    \hfill
    \begin{subfigure}{0.3\textwidth}
    \begin{centering}
   \includegraphics[width=\textwidth]{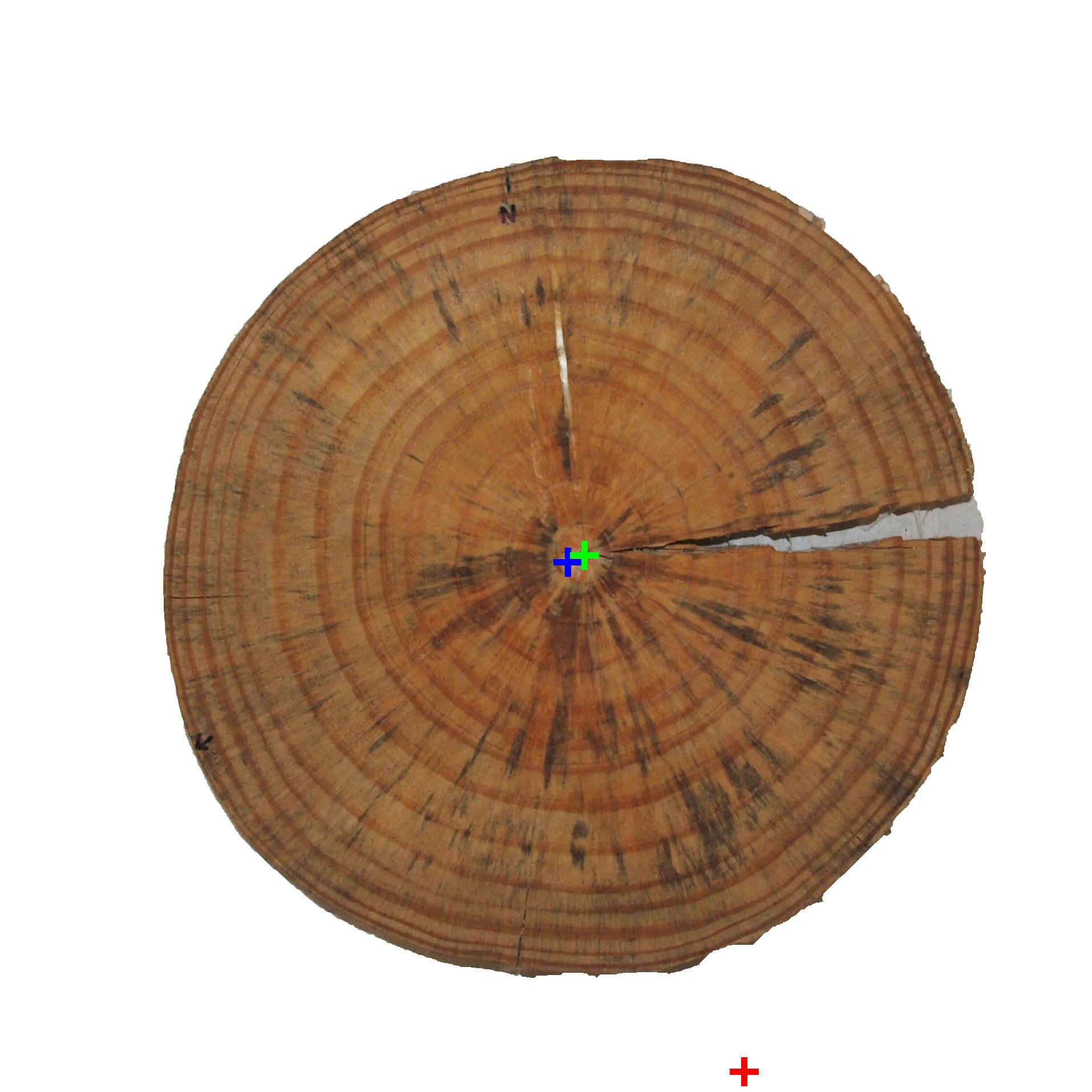}
    \label{fig:L02e_comp}
    \caption{L02e}
    \end{centering}
    \end{subfigure}
    \hfill
    \begin{subfigure}{0.3\textwidth}
    \begin{centering}
   \includegraphics[width=\textwidth]{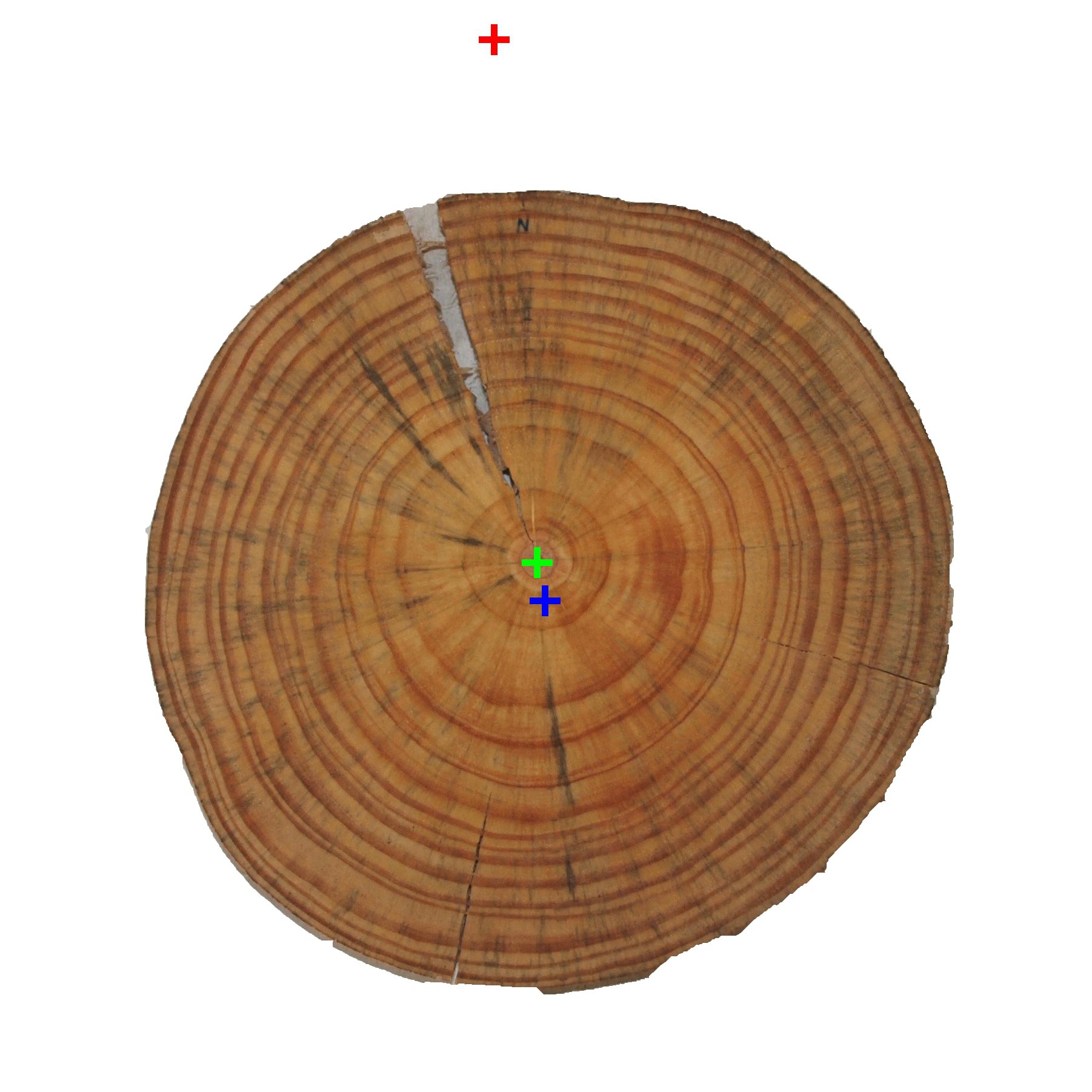}
    \label{fig:L09b_comp}
    \caption{L09b}
    \end{centering}
    \end{subfigure}
    \hfill
   \caption{Some results over the more difficult cases of the UruDendro dataset. Green: ground truth pith, red: ACO pith prediction, and blue: Shraml pith prediction.}
   \label{fig:comparison_urudendro}
\end{centering}
\end{figure}

Analyzing how both methods perform in the more difficult cases is interesting. Figure \ref{fig:comparison_urudendro} illustrates the pith prediction result over the more difficult disks in the UruDendro dataset. Despite the ACO method being more precise on average, it gives results far from the medulla region (disks L02b, L02e, and L09b) when it fails. All these disks have in common that the ring's information is very poor near the pith location. For example, disks L02b, F07e, L02a, and L02e have a black fungus over the pith location. Additionally, cracks are present in disks F04b, F07e, L02b, L02e, and L09b. Finally, in disk F04b, despite fungus and a crack, both methods perform very well (positioning the pith between rings 1 and 2).

\begin{figure}[ht]
\begin{centering}
    \begin{subfigure}{0.4\textwidth}
    \begin{centering}
   \includegraphics[width=\textwidth]{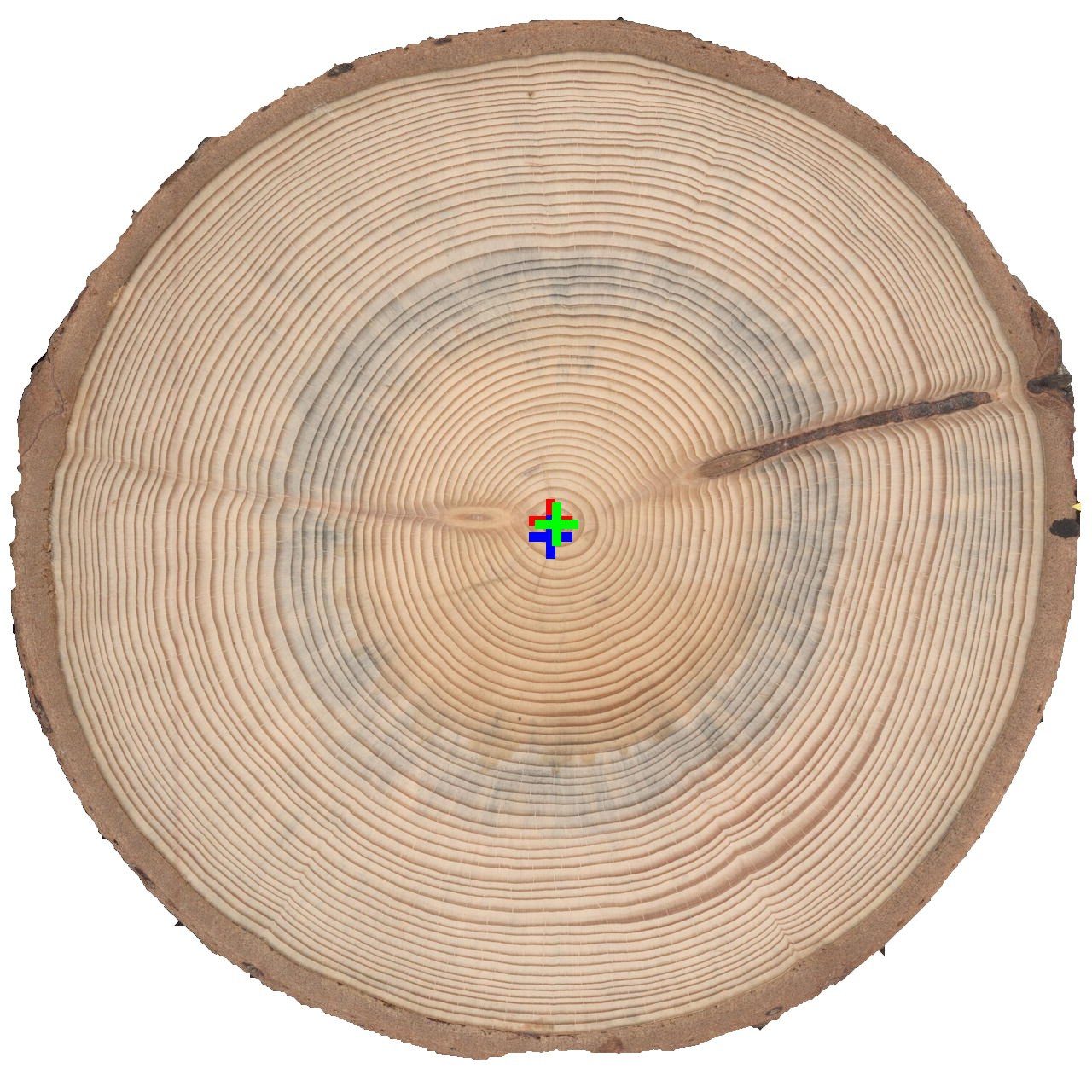}
    \label{fig:fx1_comp}
    \caption{AbiesAlba1}
    \end{centering}
    \end{subfigure}
    \hfill
    \begin{subfigure}{0.4\textwidth}
    \begin{centering}
   \includegraphics[width=\textwidth]{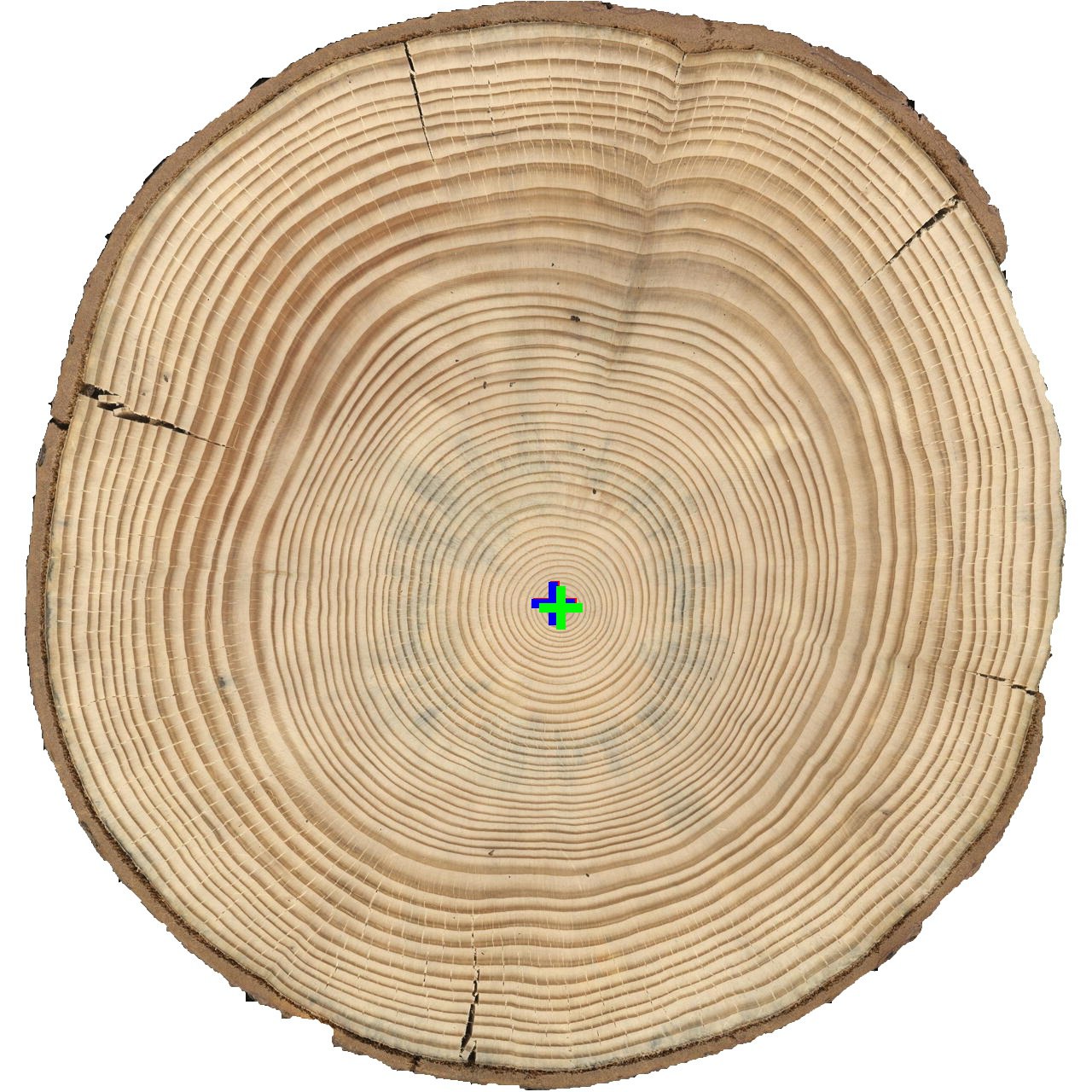}
    \label{fig:fx7_comp}
    \caption{AbiesAlba7}
    \end{centering}
    \end{subfigure}
    \hfill
   \caption{Methods detection over the most difficult cases of the Kennel dataset. Green: ground truth pith, red: ACO pith prediction, and blue: Schraml pith prediction.}
   \label{fig:comparison_kennel}
\end{centering}
\end{figure}

Figure \ref{fig:comparison_kennel} illustrates how the method performs when detection falls outside the pith region, and both methods fail in the same two disks. Despite  Table \ref{tab:kennel_results_d1} showing better indicators for the ACO method (measured in pixels), both predictions are pretty good.

\section{Conclusions}

Even though the Schraml method was published in 2013, it still gives state-of-the-art results. If ring edge information is strong (no fungus presence), the method locates the pith successfully. However, the Schraml method is not very precise if the fungus presence is heavy. 

Regarding computational time, the Schraml method makes the detection in less than 1 second (on average) for the optimal configuration\footnote{All experiments were made using a workstation with Intel Core i5 10300H and RAM 16GB.} in both datasets.

\section*{Image Credits}
\includegraphics[height=2em]{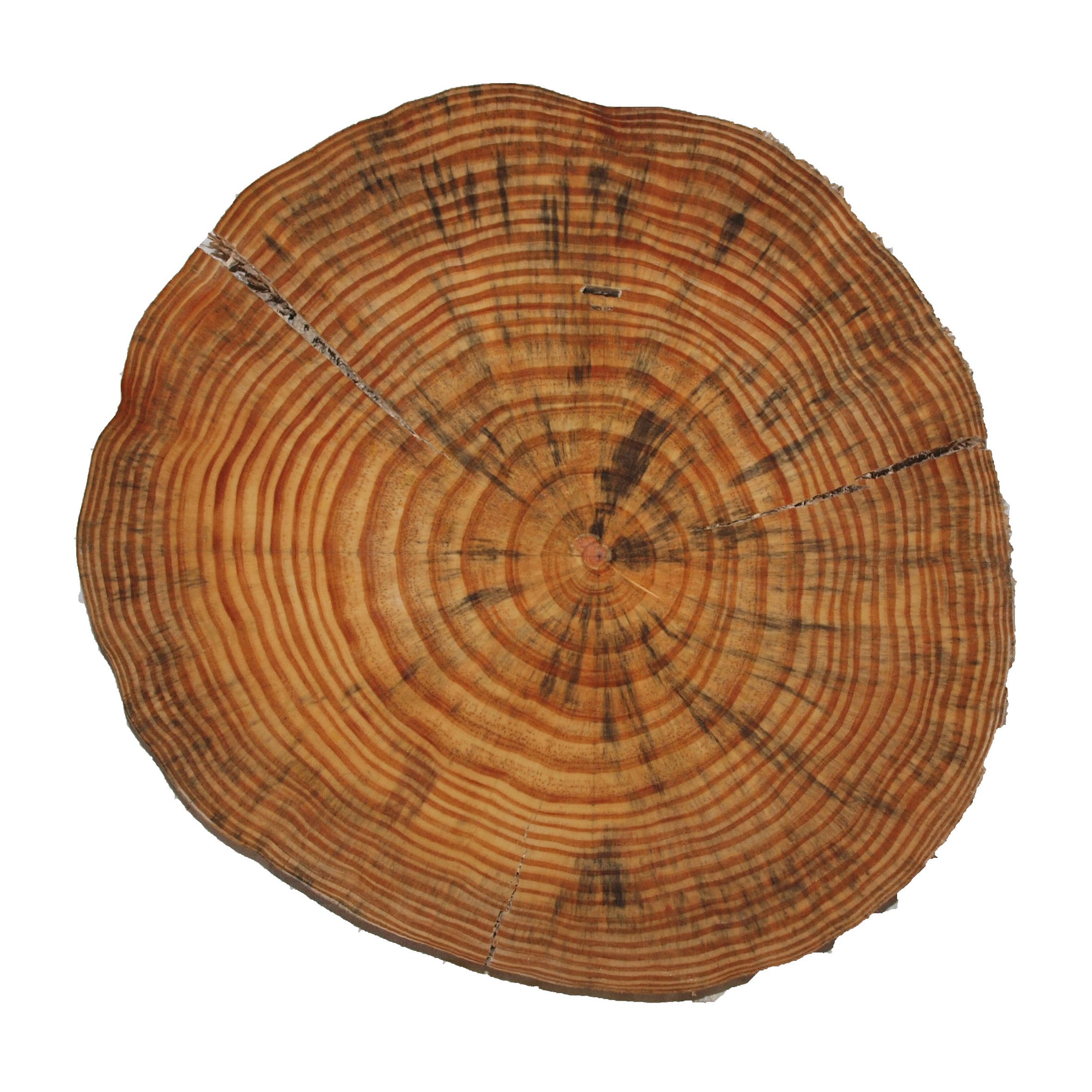}
\includegraphics[height=2em]{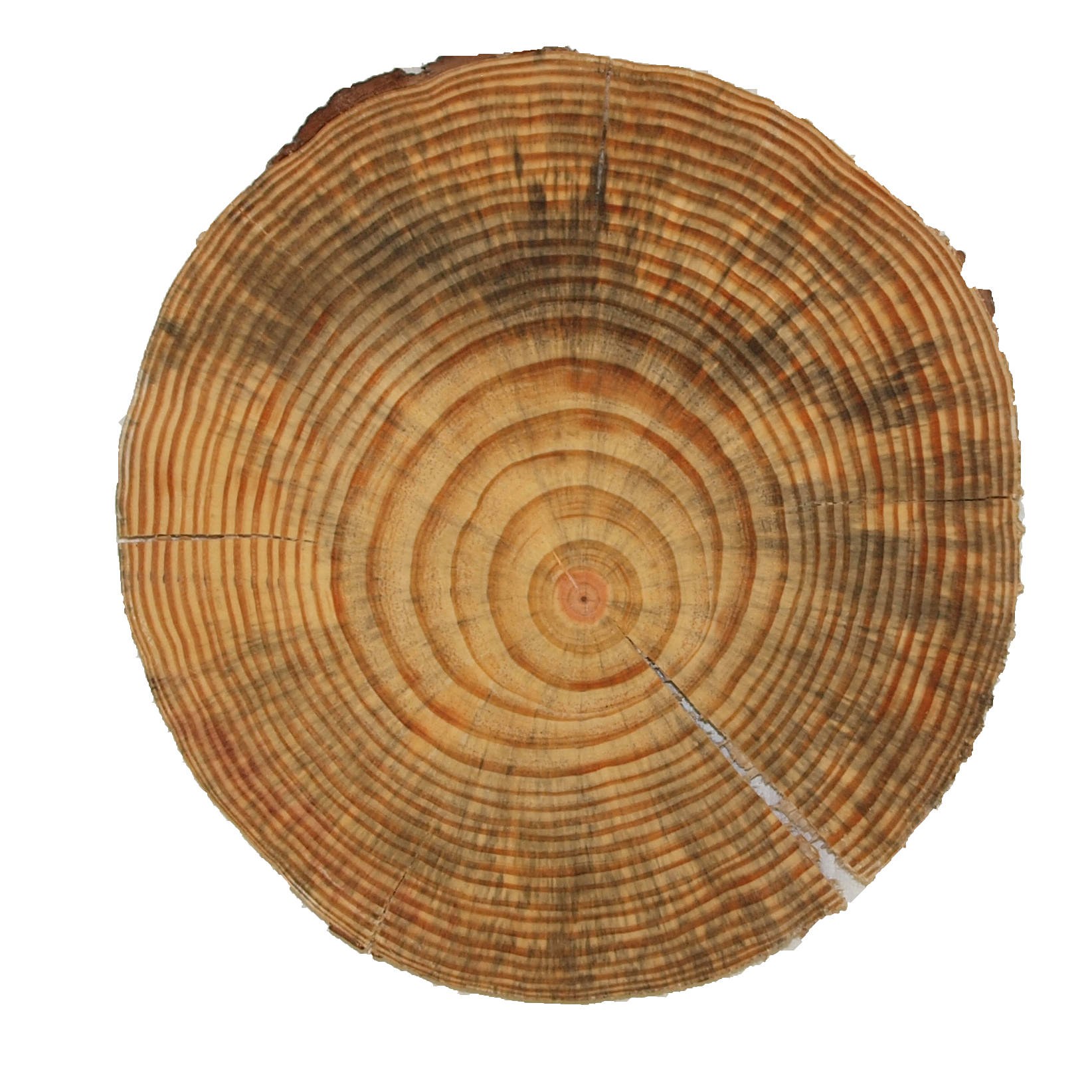}
\includegraphics[height=2em]{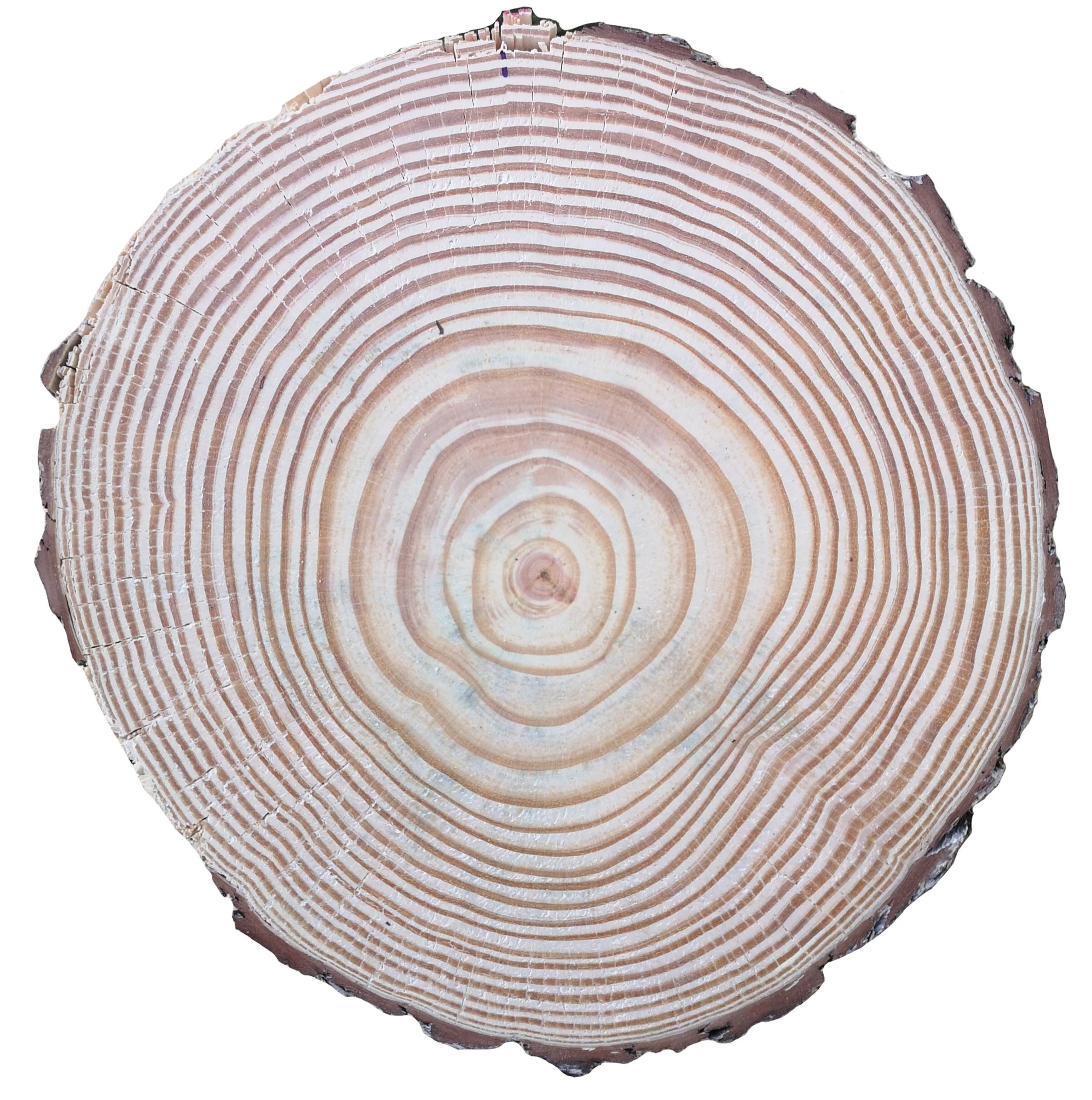}
\includegraphics[height=2em]{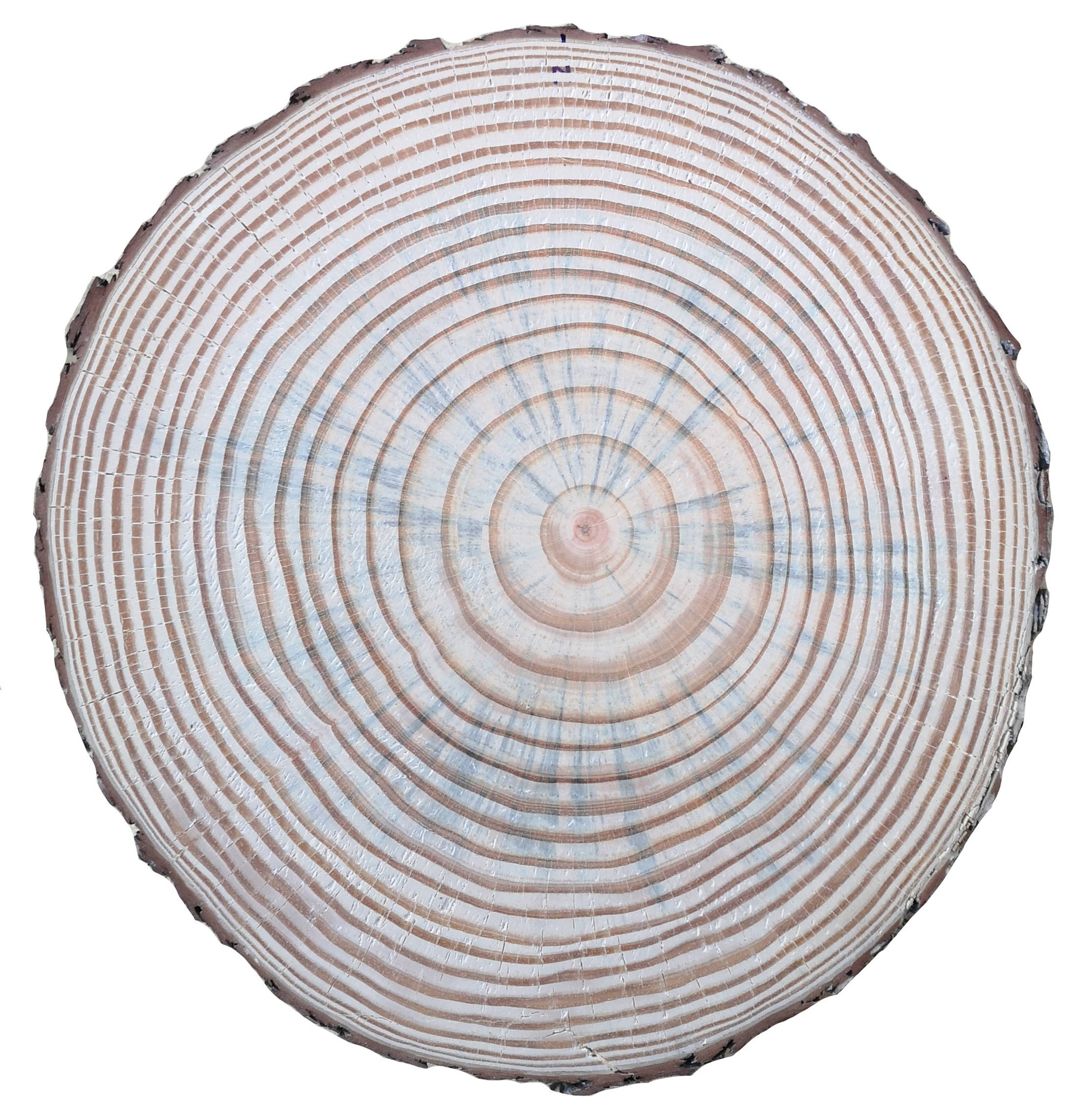}
\includegraphics[height=2em]{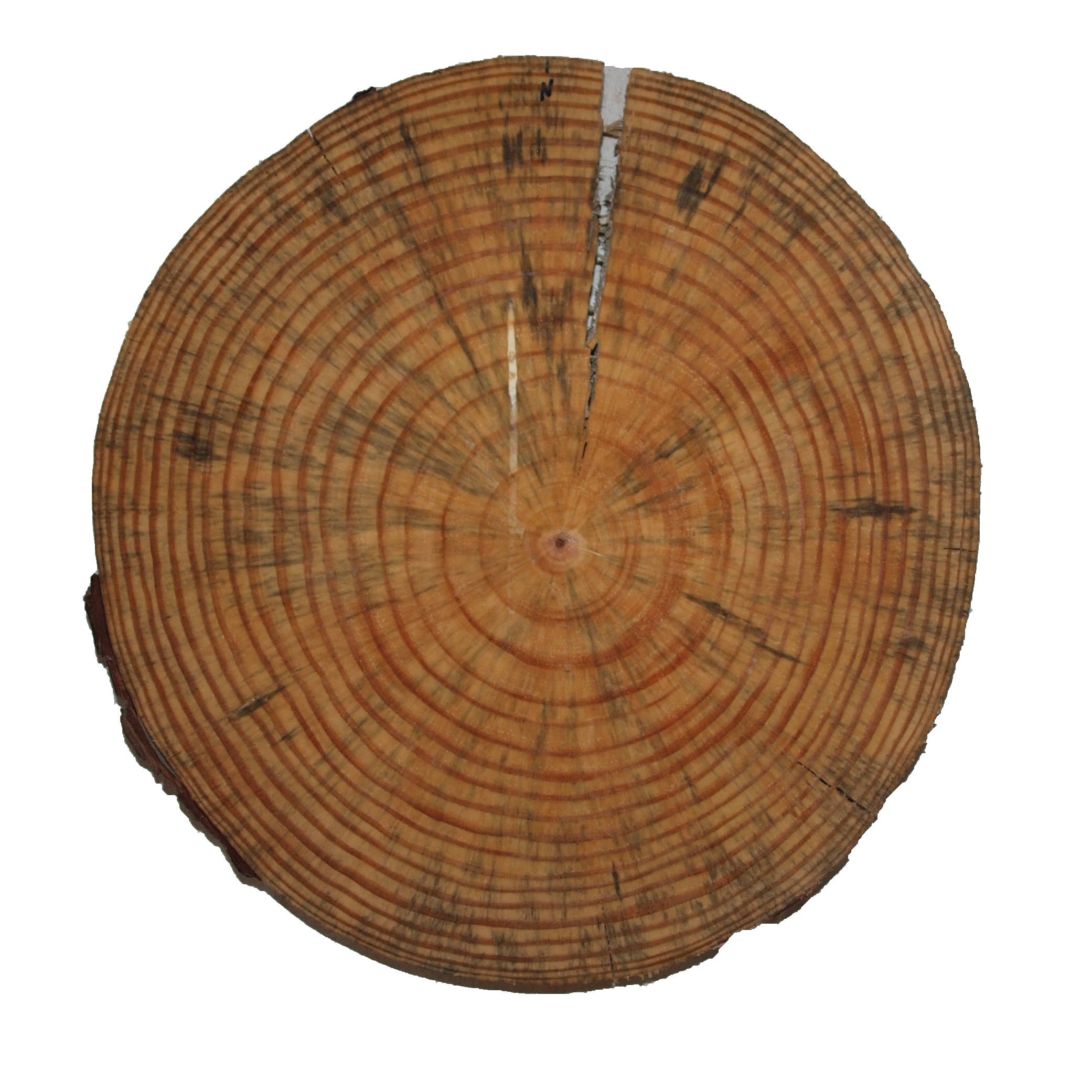}
\includegraphics[height=2em]{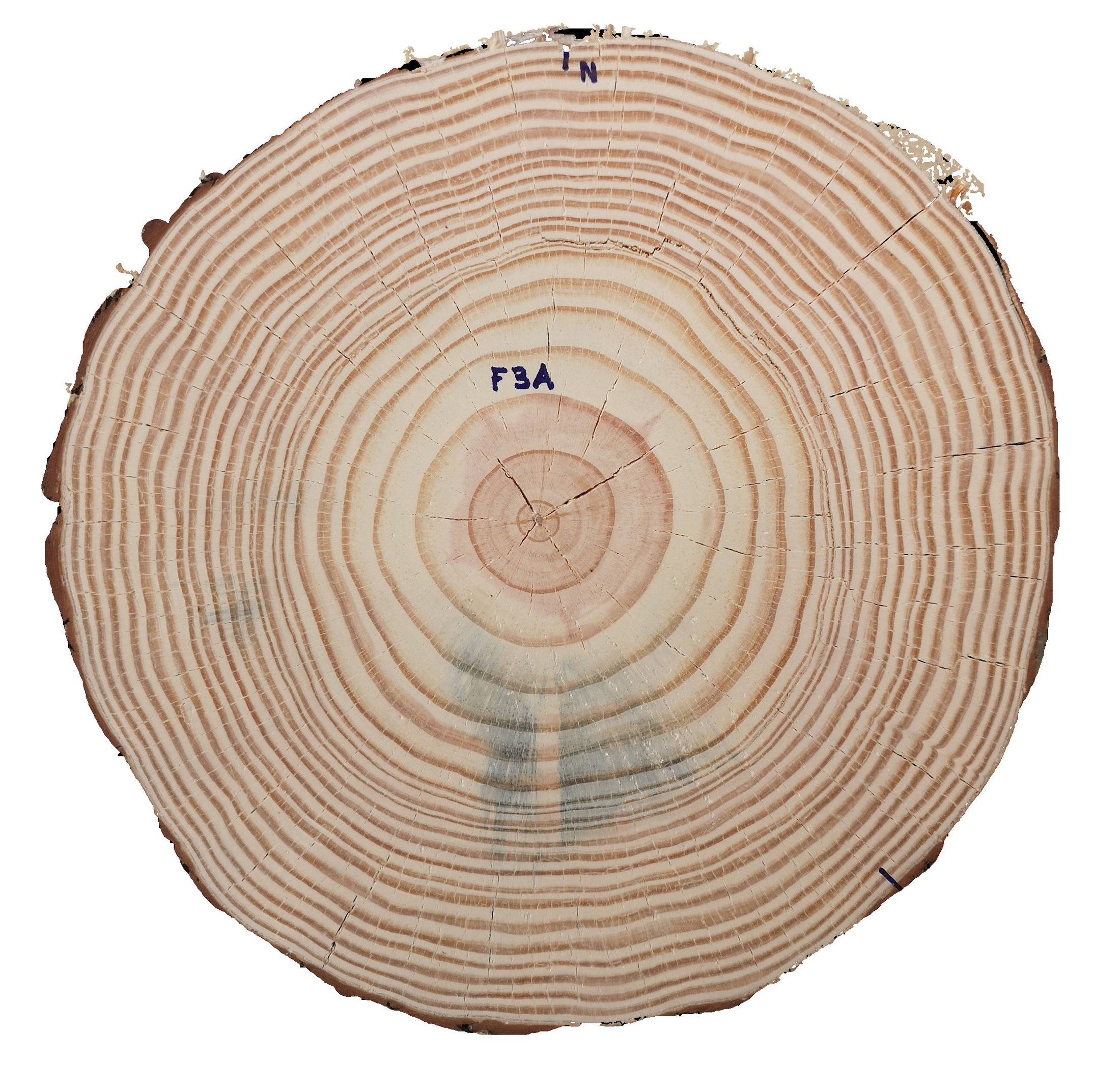}
\includegraphics[height=2em]{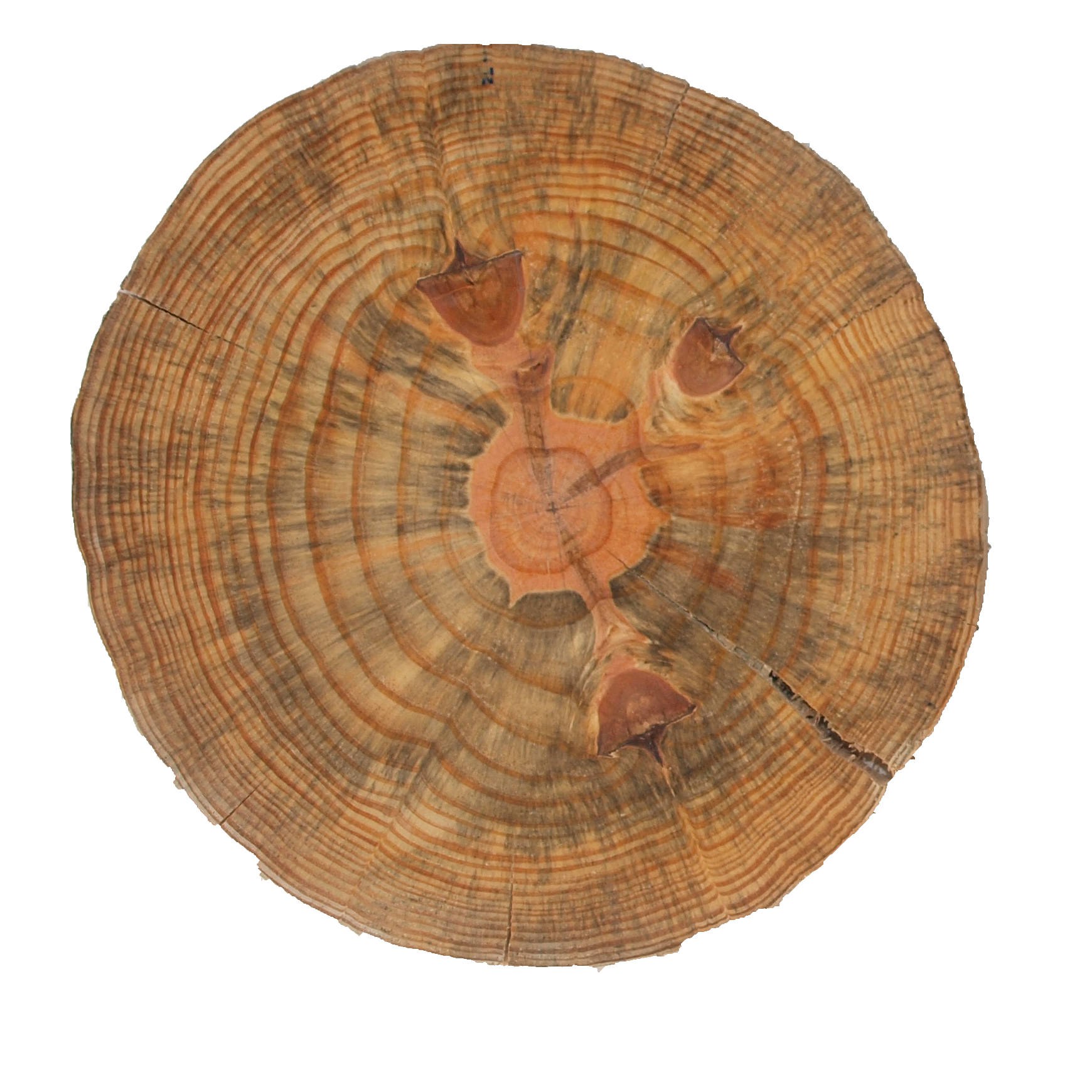}
\includegraphics[height=2em]{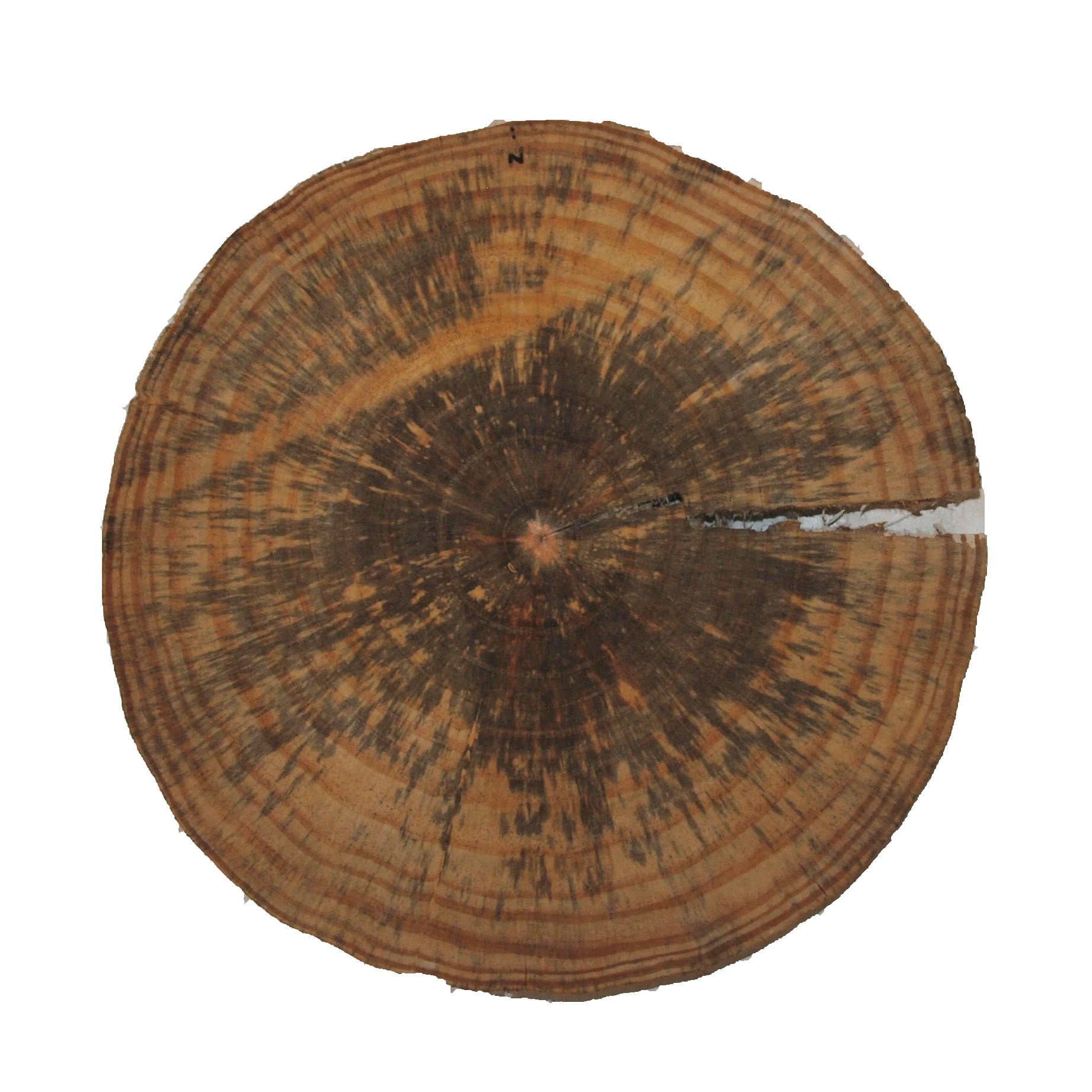}
\includegraphics[height=2em]{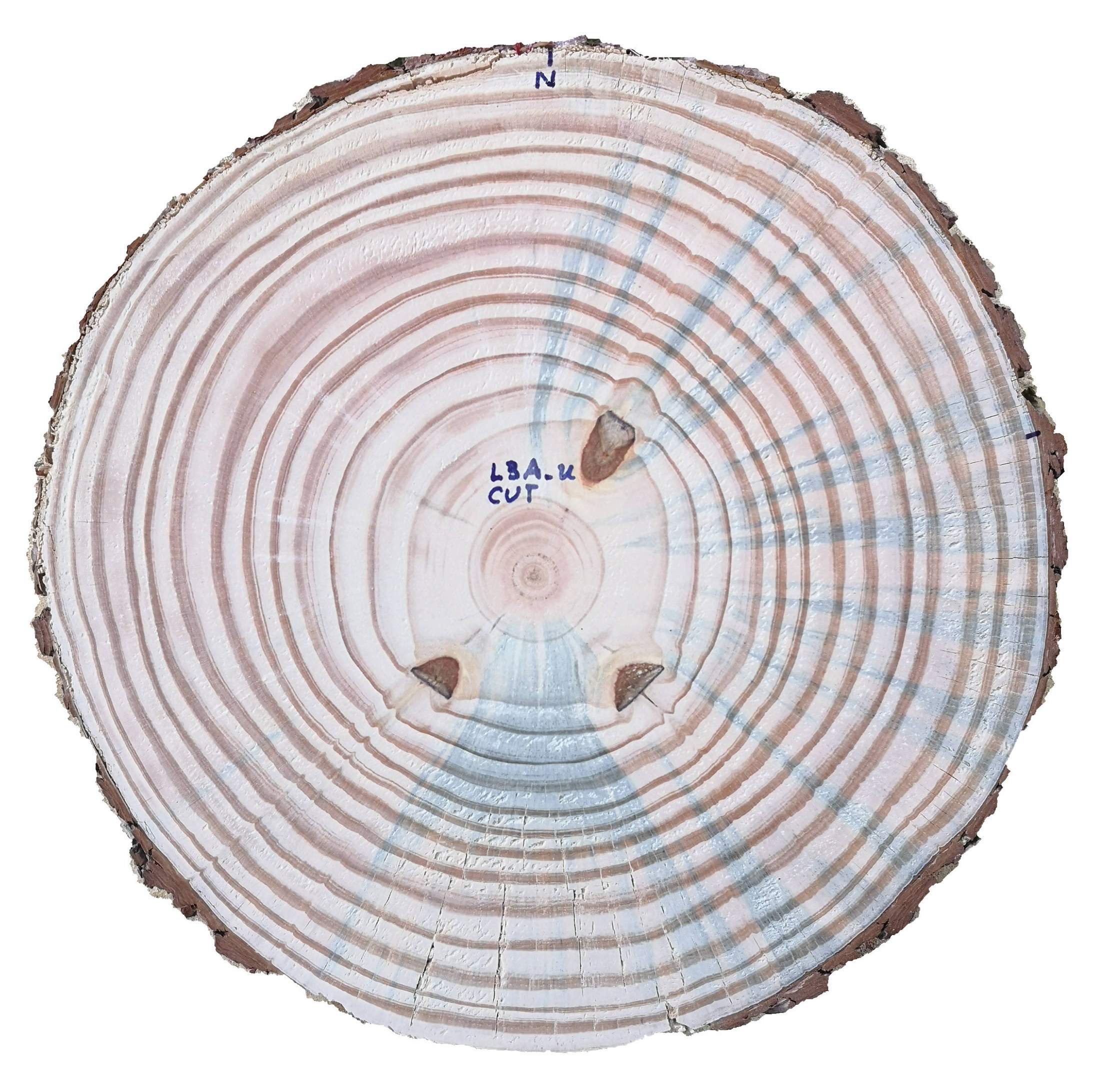}
\includegraphics[height=2em]{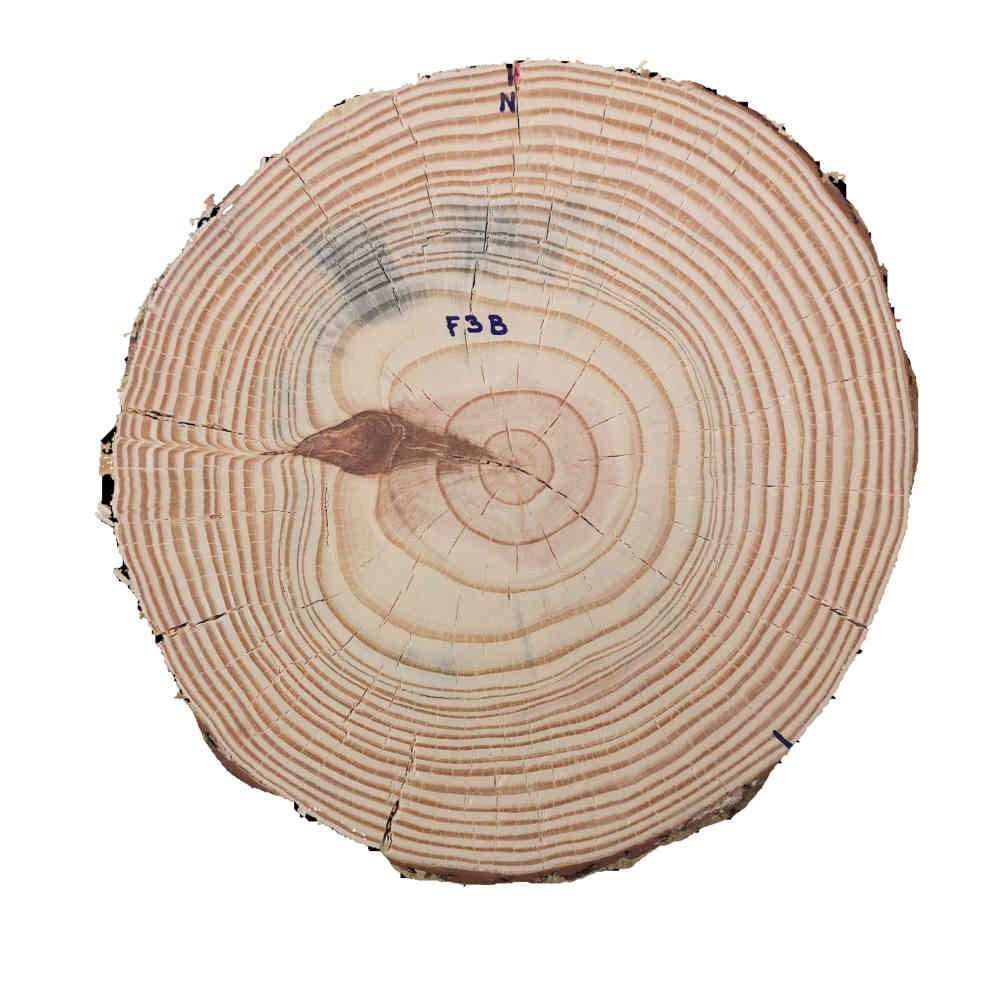}
Images from the UruDendro dataset
\\
\includegraphics[height=2em]{figures/kennel/ac1.jpg}
\includegraphics[height=2em]{figures/kennel/ac2.jpg}
\includegraphics[height=2em]{figures/kennel/ac3.jpg}
\includegraphics[height=2em]{figures/kennel/ac4.jpg}
\includegraphics[height=2em]{figures/kennel/ac5.jpg}
\includegraphics[height=2em]{figures/kennel/ac6.jpg}
\includegraphics[height=2em]{figures/kennel/ac7.jpg}
original images from the Kennel dataset.
\\


{\small
\bibliographystyle{siam}
\bibliography{bibliography}
}
\end{document}

%% file: figures/introduccion/SpyderWebb.pdf_tex
\begingroup%
  \makeatletter%
  \providecommand\color[2][]{%
    \errmessage{(Inkscape) Color is used for the text in Inkscape, but the package 'color.sty' is not loaded}%
    \renewcommand\color[2][]{}%
  }%
  \providecommand\transparent[1]{%
    \errmessage{(Inkscape) Transparency is used (non-zero) for the text in Inkscape, but the package 'transparent.sty' is not loaded}%
    \renewcommand\transparent[1]{}%
  }%
  \providecommand\rotatebox[2]{#2}%
  \newcommand*\fsize{\dimexpr\f@size pt\relax}%
  \newcommand*\lineheight[1]{\fontsize{\fsize}{#1\fsize}\selectfont}%
  \ifx\svgwidth\undefined%
    \setlength{\unitlength}{115.6778183bp}%
    \ifx\svgscale\undefined%
      \relax%
    \else%
      \setlength{\unitlength}{\unitlength * \real{\svgscale}}%
    \fi%
  \else%
    \setlength{\unitlength}{\svgwidth}%
  \fi%
  \global\let\svgwidth\undefined%
  \global\let\svgscale\undefined%
  \makeatother%
  \begin{picture}(1,0.98635745)%
    \lineheight{1}%
    \setlength\tabcolsep{0pt}%
    \put(0,0){\includegraphics[width=\unitlength,page=1]{figures/introduccion/SpyderWebb.pdf}}%
  \end{picture}%
\endgroup%

%% file: bibliography.bib
@inproceedings{Schraml2013,
  title={Pith estimation on rough log end images using local fourier spectrum analysis},
  author={Schraml, Rudolf and Uhl, Andreas},
  booktitle={Proceedings of the 14th Conference on Computer Graphics and Imaging (CGIM’13), Innsbruck, AUT},
  volume={10},
  pages={2013--797},
  year={2013},
  organization={Citeseer}
}

@book{Granlund,
author = {G. H. Granlund and H. Knutsson},
note = {ISBN~978-0-7923-9530-0},
publisher = {Kluwer Academic Publishers},
title = {Signal processing for computer vision},
year = {1995},
}

@INPROCEEDINGS{Kurdthongmee2018,
  author={Kurdthongmee, Wattanapong and Suwannarat, Korrakot and Panyuen, Praepaka and Sae-Ma, Naruedom},
  booktitle={2018 15th International Joint Conference on Computer Science and Software Engineering (JCSSE)}, 
  title={A Fast Algorithm to Approximate the Pith Location of Rubberwood Timber from a Normal Camera Image}, 
  year={2018},
  volume={},
  number={},
  pages={1-6},
  doi={10.1109/JCSSE.2018.8457375}}

@article{Decelle2022,
    title   = {{Ant Colony Optimization for Estimating Pith Position on Images of Tree Log Ends}},
    author  = {Decelle, Rémi  and Ngo, Phuc and Debled-Rennesson, Isabelle and Mothe, Frédéric and Longuetaud, Fleur},
    journal = {{Image Processing On Line}},
    volume  = {12},
    pages   = {558--581},
    year    = {2022},
    note    = {\url{https://doi.org/10.5201/ipol.2022.338}}
}

@article{Gazo2020,
title = {{A fast pith detection for computed tomography scanned hardwood logs}},
journal = {Computers and Electronics in Agriculture},
volume = {170},
pages = {105107},
year = {2020},
issn = {0168-1699},
doi = {https://doi.org/10.1016/j.compag.2019.105107},
url = {https://www.sciencedirect.com/science/article/pii/S0168169919307598},
author = {Rado Gazo and Juraj Vanek and Michel Abdul\_Massih and Bedrich Benes}
}

@article{KURDTHONGMEE2020,
title = {A comparative study of the effectiveness of using popular DNN object detection algorithms for pith detection in cross-sectional images of parawood},
journal = {Heliyon},
volume = {6},
number = {2},
pages = {e03480},
year = {2020},
issn = {2405-8440},
doi = {https://doi.org/10.1016/j.heliyon.2020.e03480},
url = {https://www.sciencedirect.com/science/article/pii/S240584402030325X},
author = {Wattanapong Kurdthongmee}
}

@article{UruDendro,
  author = {Marichal, Henry and Passarella, Diego and Lucas, Christine and Profumo, Ludmila and Casaravilla, Veronica and Rocha Galli, Maria Noel and Ambite, Serrana and Randall, Gregory},
  title = {UruDendro, a public dataset of 64 cross-section images and manual annual ring delineations of {Pinus taeda} {L}.},
  journal = {Annals of Forest Science},
  volume = {82},
  pages = {25},
  year = {2025},
  doi = {10.1186/s13595-025-01296-5},
  url = {https://doi.org/10.1186/s13595-025-01296-5}
}

@article{KennelBS15,
  author    = {Pol Kennel and
               Philippe Borianne and
               G{\'{e}}rard Subsol},
  title     = {An automated method for tree-ring delineation based on active contours
               guided by {DT-CWT} complex coefficients in photographic images: Application
               to Abies alba wood slice images},
  journal   = {Comput. Electron. Agric.},
  volume    = {118},
  pages     = {204--214},
  year      = {2015},
  url       = {https://doi.org/10.1016/j.compag.2015.09.009},
  doi       = {10.1016/j.compag.2015.09.009},
  bibsource = {dblp computer science bibliography, https://dblp.org}
}
